\documentclass[12pt,a4paper]{article}

\usepackage[english]{babel}
\usepackage[utf8]{inputenc}
\usepackage[T1]{fontenc}
\usepackage{lmodern}
\usepackage{geometry}
\usepackage{setspace}
\usepackage{hyperref}
\usepackage{graphicx}
\usepackage{booktabs}
\usepackage{parskip}
\usepackage{subcaption}
\usepackage{amsmath}
\usepackage{float}
\usepackage{csquotes}
\usepackage{longtable}
\usepackage{pdfpages}

\usepackage[style=numeric-comp, sortcites=true, sorting=none]{biblatex}
\AtEveryBibitem{
  \clearfield{urlyear}
  \clearfield{urlmonth}
  \clearfield{urlday}
}

\begin{document}

\begin{titlepage}
	\thispagestyle{empty}
	
	\begin{flushleft}
		
		{\Large\bfseries Carl von Ossietzky \\
			Universität Oldenburg}
		
		\vspace{1.5cm}
		
		{\large\bfseries Fach-Bachelor Informatik}\\
		
		\vspace{2cm}
		
		{\large\bfseries BACHELORARBEIT}
		
		\vspace{1.5cm}
		
		{\large\bfseries
			ECG for laboratory values:\\
			A comprehensive physiological and technical analysis
		}
		
		\vspace{3cm}
		
		{\bfseries vorgelegt von} \\
		Walentin Palanchuk
		
		\vfill
		
		{\bfseries Betreuender Gutachter}\\
		Prof. Dr. Nils Strodthoff
		
		\vspace{0.5cm}
		
		{\bfseries Zweiter Gutachter}\\
		Dr. Juan Miguel Lopez Alcaraz
		
		\vspace{1.5cm}
		
		{\bfseries Oldenburg, 31.08.2026}
		
	\end{flushleft}
	
\end{titlepage}
\section*{Abstract}
Laboratory measurements provide important information in clinical decision making but typically require invasive methods, such as blood sampling. Recent research suggests that ECG recordings may contain information beyond cardiac rhythm analysis and may provide information about laboratory values. This thesis investigates the prediction of N-Terminal Pro-B-Type Natriuretic Peptide (NT-proBNP) from ECG recordings and patient metadata using a multimodal deep learning framework, while considering the physiological plausibility of ECG-based NT-proBNP prediction. Furthermore, this work investigates the influence of a 1-lead configuration using lead II compared to the corresponding 12-lead ECG configuration on predictive performance. Additional experiments on glucose, hemoglobin and potassium are presented in the appendix.

Regression, binary classification and three-class classification experiments were conducted using the MIMIC-IV database. The three-class classification is based on clinically established thresholds. This work showed that the 12-lead configurations achieved consistently better predictive performance. The regression experiments showed a mean absolute error of 534.98~pg/mL for the 1-lead configuration and 514.27~pg/mL for the 12-lead configuration. The classification experiment achieved an AUROC of 0.849 for the 1-lead binary classification and 0.876 for the 12-lead configuration. The three-class classification achieved macro-averaged AUROC values of 0.738 and 0.752 for the 1-lead and 12-lead configurations respectively. The regression experiments showed that the precise prediction of NT-proBNP remains challenging, while the classification experiments showed that broader concentration ranges may be classified.

The findings are consistent with the physiological hypothesis that ECG characteristics and NT-proBNP concentrations may reflect shared underlying cardiovascular processes. Consequently, the ECG-based NT-proBNP prediction appears to be feasible as a complementary source of information, particularly when classification of broader concentration ranges is needed.
However, laboratory measurements remain necessary and further research using independent datasets is required.

\newpage
\tableofcontents

\clearpage
\section{Introduction}
This chapter introduces the motivation of ECG-based laboratory value prediction and the clinical need for noninvasive approaches to obtain information related to laboratory values. Furthermore, this chapter discusses the role of machine learning in this context, the existing research and the contributions of this work.
\subsection{Clinical Need}
Laboratory values have a crucial role in modern healthcare and are indispensable for clinical decision making. They provide information on the health status of the patient and are routinely used for disease diagnosis, treatment planning and monitoring the effectiveness of therapies\cite{ferraroLABORATORY2016}. Biomarkers that are measured through laboratory testing, such as NT-proBNP, hemoglobin, potassium and glucose, support clinicians in evaluating a wide range of cardiovascular, metabolic and physiological conditions\cite{mckieNTPROBNP2016, petersenREGULATION2017}.

Despite their clinical importance, many laboratory values require invasive methods, such as blood sampling. Although blood sampling is invasive, it remains one of the most widely used testing methods\cite{lima-oliveiraLABORATORY2015}. It requires trained personnel, laboratory infrastructure and processing time for the results to become available. This contributes to the costs associated with laboratory testing for both patients and healthcare providers. Furthermore, repeated blood sampling increases workload and may cause discomfort for the patients, especially when frequent monitoring is required\cite{yeshouaINTERVENTI2023}. Hence, an approach that can provide clinically relevant information faster and causes less discomfort for patients is in demand.
\subsection{Machine Learning in Healthcare}
Machine learning has become a substantial field in modern healthcare, demonstrating potential to support clinical decision making and improve patient care. Prior research has shown that it may support clinicians by providing fast and accurate insights about patient information and assisting clinical decision making\cite{alcarazhancingclinicaldecision2025}.
Common applications of machine learning in healthcare include image analysis and analysis of biological and clinical data\cite{habehhMACHINELE2021, patelMACHINELE2015}. 

ECG signals have become an important data source for machine learning. Recent research suggests that ECG contains information beyond cardiac rhythm analysis and may reflect broader physiological and pathological processes\cite{alcaraznormalitypredictionforecasting2025}. Since ECG signals can be obtained easily and without invasive procedures using wearable devices such as smartwatches, they represent a promising data source to be used with machine learning. 
Furthermore, multimodal machine learning is used by combining multiple data sources, such as physiological signals and patient metadata. Multimodal machine learning is therefore more suitable for use in the medical context, because it mimics the multimodal decision making of clinical experts\cite{klineMULTIMODAL2022}. These developments have motivated the application of multimodal deep learning for predicting laboratory values from ECG signals and patient metadata.
\subsection{Research Gap}
The prediction of laboratory values from electrocardiography has become an increasingly active area of research in recent years. Previous studies have shown that ECG signals contain information beyond traditional cardiac rhythm analysis\cite{ceasovschihELECTROCAR2024, marzoogMACHINELE2026}. In addition, several studies have investigated the prediction of laboratory values using deep learning approaches\cite{alcaraznormalitypredictionforecasting2025, gallowayDEVELOPMEN2019}, demonstrating that clinically relevant information may be extracted from ECG signals.

Despite these results, existing research has focused primarily on the predictive performance of machine learning approaches, while the physiological plausibility of ECG-based laboratory value prediction and the influence of technical design choices on predictive performance have been less extensively studied. 

These limitations motivate a more comprehensive investigation of ECG-based laboratory value prediction that considers both physiological plausibility and technical aspects in addition to predictive performance. 
\subsection{Contributions}
This thesis investigates ECG-based laboratory value prediction with NT-proBNP as the primary laboratory biomarker. In addition to predictive performance, this work considers the physiological relationships and technical design choices that may influence ECG-based prediction. The main contributions of this thesis are:
\begin{itemize}
    \item Evaluation of NT-proBNP prediction using a multimodal deep learning approach combining ECG recordings and patient metadata.
    \item Comparison of 1-lead and 12-lead ECG configurations across regression and classification tasks.
    \item Physiological and technical analysis of ECG-based NT-proBNP prediction.
    \item Additional experiments on hemoglobin, glucose, potassium and further NT-proBNP configurations.
\end{itemize}
\clearpage
\section{Background and Theoretical Foundations}
This chapter aims to provide a theoretical background required for the investigation of ECG-based laboratory prediction. It introduces the physiological foundation of electrocardiography and NT-proBNP and briefly describes the deep learning concepts and model architectures used in this work. 
\subsection{Electrocardiography}
The electrocardiogram is a noninvasive test that records the electrical activity of the heart throughout a cardiac cycle by using electrodes attached to the skin of the patient. It is commonly used to detect heart rhythm problems and is essential in medical care. Furthermore, recent studies suggest that ECG signals may contain information to help detect noncardiac diseases such as pancreatitis or strokes\cite{ceasovschihELECTROCAR2024}, showing that ECG may reflect broader physiological and pathological processes beyond traditional cardiac rhythm analysis. A typical clinical ECG consists of twelve leads, each providing a different perspective of the electrical activity of the heart.

An electrocardiogram is based on the indirect recording of the dipole vectors of depolarization and repolarization of cardiomyocytes, the muscle cells that make up the heart, over time. During the cardiac cycle, the heart contracts due to action potentials traveling through the chambers of the heart. During this, generally one part of the cardiac tissue is depolarized, while another is at resting potential. Consequently, a dipole is created, which causes current flow in surrounding body fluids, resulting in fluctuating electric fields that are detected by the electrodes on the patient skin\cite{iaizzoHANDBOOKC2024}.

The resulting ECG waveforms consist of several characteristic waves, segments and intervals that summarize electrical events during a cardiac cycle. The appearance of these waveform components depends on the position and orientation of the recording electrodes relative to the direction of cardiac electrical activity. Thus, different ECG leads provide complementary information about the heart and may vary in their ability to capture physiological and pathological changes\cite{iaizzoHANDBOOKC2024}.
\subsubsection{ECG Waveform Components}
A standard ECG waveform consists of several characteristic components that represent different electrical events during the cardiac cycle. The P wave represents atrial depolarization. The QRS complex represents the electrical depolarization of the ventricles, while the T wave represents the ventricular repolarization\cite{buxtonACCAHAHR2006}.

In addition to the individual waves, intervals and segments are used in the ECG to describe the timing of cardiac electrical activity\cite{buxtonACCAHAHR2006}. The PR interval describes the time from the beginning of atrial depolarization to the beginning of ventricular depolarization, while the duration of the ventricular depolarization is described by the QRS duration. The QT interval represents the duration of ventricular depolarization and repolarization, while the RR interval reflects the time between two successive ventricular depolarizations\cite{buxtonACCAHAHR2006}.

Changes in the morphology, duration or amplitude of components in the ECG reflect alterations in cardiac electrical activity\cite{surawiczAHAACCFH2009}. These changes may be associated with pathological or physiological processes\cite{surawiczAHAACCFH2009}. Consequently, the ECG waveform contains information that may indicate changes in the cardiovascular state of a patient and may contain patterns that can be detected by machine learning models\cite{alcaraznormalitypredictionforecasting2025}.
\subsubsection{ECG Lead Configuration}
An ECG lead shows the electrical activity of the heart from a specific spatial perspective\cite{kligfieldRECOMMENDA2007}. Since cardiac electrical activity propagates through the heart in different directions, the waveform can vary depending on the corresponding ECG lead orientation\cite{kligfieldRECOMMENDA2007}. Therefore, different ECG leads provide complementary views of the same cardiac electrical activity and certain waveform characteristics may be more apparent in some lead configurations compared to others\cite{kligfieldRECOMMENDA2007}.

A typical 12-lead ECG consists of six limb leads and six precordial leads. The limb leads include lead I, II, III and the augmented leads aVR, aVL and aVF. The precordial leads include V1 to V6 and are recorded from different positions on the chest\cite{kligfieldRECOMMENDA2007}. These leads provide multiple angles of cardiac electrical activity and allow electrical changes in different regions of the heart to be captured\cite{kligfieldRECOMMENDA2007}.

Lead II records the electrical activity between the electrode on the right arm and the electrode on the left leg\cite{kligfieldRECOMMENDA2007}. Due to its electrode orientation, it commonly provides clearly distinguishable P waves, QRS complexes and T waves\cite{issaHEARTBEAT2023} and is often used for rhythm monitoring and 1-lead ECG recordings\cite{issaHEARTBEAT2023}.
\subsection{NT-proBNP}
N-Terminal Pro-B-Type Natriuretic Peptide is the main laboratory biomarker investigated in this thesis. It is part of the natriuretic peptide system of the heart, which regulates blood pressure and fluid balance in response to cardiovascular stress\cite{iaizzoHANDBOOKC2024}. In response to increased myocardial wall stress, cardiac muscle cells produce the precursor molecule proBNP\cite{nishikimiBTYPENATR2022}, which is then cleaved into the biologically active hormone BNP and the biologically inactive NT-proBNP\cite{yandleBTYPENATR2015}.

NT-proBNP is widely used as a biomarker for clinical assessment of heart failure. Low concentrations help to exclude heart failure, whereas elevated concentrations may help with diagnosis and may provide information about the severity of the disease\cite{oremusSYSTEMATIC2014, minamisawa2023UPDAT2025}. However, elevated NT-proBNP concentrations are not specific to a single cardiac condition. Therefore, additional clinical information should be considered when interpreting NT-proBNP concentrations\cite{baggishDIFFERENTI2008}. 

NT-proBNP concentrations may also be influenced by patient-specific and physiological factors. Age, renal function and cardiac rhythm have been associated with NT-proBNP concentrations\cite{COMPARISON}. Therefore, the actual NT-proBNP concentration may vary between patients depending on their individual clinical characteristics, which need to be considered when NT-proBNP measurements are clinically interpreted\cite{bayes-genisPRACTICAL2023}.

Clinical interpretation of NT-proBNP commonly involves concentration thresholds that assist in the assessment of suspected heart failure\cite{OVERVIEWC2018}. The National Institute for Health and Care Excellence uses thresholds of 400~pg/mL and 2000~pg/mL to distinguish levels of urgency for further assessment\cite{OVERVIEWC2018}, while in acute settings, an NT-proBNP concentration threshold of 300~pg/mL is commonly used to rule out acute heart failure\cite{januzziNTPROBNPT2006}. This illustrates that NT-proBNP is interpreted both as a precise measurement and with clinically relevant concentration ranges.
\subsection{ECG Characteristics and NT-proBNP}
Despite the fact that ECG recordings and NT-proBNP measurements represent different aspects of cardiac function, both can reflect changes associated with cardiovascular disease\cite{neyaziDEEPLEARN2024}. Furthermore, elevated NT-proBNP concentrations have been associated with ECG characteristics and abnormalities related to cardiac dysfunction\cite{abdinEFFECTSDA2024}. This suggests that physiological processes associated with changes in NT-proBNP concentration may also be reflected in the ECG recording.

Recent studies have further investigated the relationship between ECG recordings and NT-proBNP concentrations using machine learning, more specifically deep learning. Indolfi et al. developed an AI-based deep learning approach using 12-lead ECG recordings to estimate elevated NT-proBNP concentrations and reported good predictive performance across multiple NT-proBNP thresholds\cite{indolfiDEVELOPMEN2026}. This further supports the hypothesis that ECG recordings contain information associated with NT-proBNP concentration.
\subsection{Deep Learning}
The clinical interpretation of electrocardiograms requires specialized knowledge and is time intensive, especially when large numbers of recordings must be evaluated\cite{prajapatiCOMPREHENS2026}. Due to this, automated interpretation algorithms have been integrated into ECG machines to support clinicians. However, their diagnostic performance remains limited for clinical use, due to the inaccuracy of these algorithms\cite{sauEMERGIN2023, kraikMOSTCOMMO2025}. Relying on automatic ECG analysis may lead to serious consequences such as delayed treatment\cite{rafieECGINTERP2021}, which has motivated considerable research into artificial intelligence, and more specifically deep learning, to support automated electrocardiogram interpretation\cite{martinez-sellesCURRENTFU2023}.

Deep learning is a subfield of machine learning that uses multiple processing layers to learn representations of data and is inspired by the way the brain processes data\cite{sarkerDEEPLEARN2021}. The individual processing layers transform the input data into representations at various abstraction levels, hence deep learning models automatically identify patterns and features which are relevant to a specific task. This reduces the need to define relevant features manually before training the model\cite{lecunDEEPLEARN2015}.

In supervised deep learning, which is used in this work, a model is trained using input data together with corresponding target values. The difference between the predicted and actual output is quantified by a loss function. Through backpropagation, this error is propagated through the network and used iteratively to adjust the parameters\cite{lecunDEEPLEARN2015}.
\subsubsection{Structured State Space Models}
Structured State Space Sequence models (S4) are neural network architectures used for processing sequential data and are based on the concept of state space models, which represent a sequence through an internal state that changes when new information is received\cite{guEFFICIENTL2022}. This allows information from previous parts of the sequence to influence the processing of later parts and therefore enables temporal dependencies within sequential data to be modeled.

This is particularly relevant for time series data, such as ECG signals, where individual samples are temporally related and patterns may extend across multiple time steps. S4 can model dependencies between different parts of an input sequence\cite{guEFFICIENTL2022}. This allows for the model to learn representations that incorporate temporal information from different parts of the ECG input\cite{mehariADVANCING2022}.

A challenge of sequence models is the efficient processing of long input sequences. S4 was developed to address this challenge by using structured parameterization of the state space model for sequence operations to be computed more efficiently\cite{guEFFICIENTL2022}. S4 can then represent the state space operation as a convolution over the input sequence, which enables efficient parallel processing during training\cite{guEFFICIENTL2022}.

Furthermore, recent work has investigated the performance of different deep learning architectures for ECG analysis. Al-Masud et al. compared multiple ECG models across a broad range of clinical prediction tasks and reported considerable differences between model architectures\cite{al-masudBENCHMARKI2026}. This further supports the use of structured state-space approaches for ECG-based prediction tasks.
\subsubsection{RealMLP}
Another neural network architecture is RealMLP, which is used for tabular data. Tabular data consists of structured data that are typically organized in rows and columns, where each column represents a specific feature. In clinical datasets, these data include demographic information, physiological measurements and other patient information.

RealMLP is an improved multilayer perceptron developed for regression and classification tasks on tabular data and processes input features through multiple connected layers that can be used for prediction like standard multilayer perceptrons\cite{holzmullerBETTERDEF2024}. The difference from regular multilayer perceptrons is that RealMLP is optimized for tabular data through specialized preprocessing, architecture and optimizations\cite{holzmullerBETTERDEF2024}.
\subsubsection{Multimodal Deep Learning}
In this work, multimodal deep learning was used to combine information from multiple modalities and potentially improve predictive results. This type of deep learning is commonly used in clinical machine learning because clinical data are often heterogeneous\cite{huddarPREDICTING2016}. Furthermore, the use of multiple different modalities may support the model by providing complementary information which would otherwise not be available from a single modality alone\cite{kronesREVIEWMUL2025}.

Since the modalities differ, separate encoders are commonly utilized. Encoders convert raw input data into usable representations for the following prediction model\cite{stahlschmidtMULTIMODAL2022}. Subsequently, the encoded representations need to be combined using a fusion strategy. Fusion strategies are classified as early fusion, intermediate fusion or late fusion.

In early fusion, modalities are combined at or near the input level and then passed to the prediction model. In intermediate fusion, both modalities are processed into separate representations, combined within the model, and then used for prediction. In late fusion, each modality is processed by independent sub-models into separate predictions which are then combined\cite{stahlschmidtMULTIMODAL2022}.
\clearpage
\section{Related Work}
Previous research has investigated the use of ECG recordings for various deep learning applications. The following sections review the ECG-based disease detection and prediction of laboratory values from ECG recordings.
\subsection{ECG-Based Disease Detection}
Deep learning has been used for various ECG-based disease detection and classification tasks\cite{xiaoDEEPLEARN2023}. A major area of early research focused primarily on automated identification of arrhythmia with positive results\cite{xiaoDEEPLEARN2023}. A deep neural network trained on 1-lead ECG signals achieved performance results similar to those of certified cardiologists across multiple rhythm classes\cite{hannunCARDIOLOGI2019}, showing that clinically relevant rhythm patterns may be learned directly from ECG signals.

Deep learning architectures have also been evaluated on large publicly available 12-lead ECG datasets, such as PTB-XL\cite{strodthoffDEEPLEARN2021}. Experiments on the PTB-XL dataset compared time series classification algorithms for different ECG prediction tasks. These experiments showed that convolutional neural networks had the strongest performance throughout the investigated tasks\cite{strodthoffDEEPLEARN2021}.

Beyond rhythm and morphology classification, ECG-based deep learning has also been used for noncardiac disease detection. A multimodal approach combining ECG and photoplethysmography signals was used to distinguish stroke patients from healthy elderly patients\cite{yuAIBASEDST2022}. Studies such as this indicated that ECG-derived information may be used for noncardiac diagnostic tasks, especially when combined with an additional modality\cite{rathnayakePHYSIOLOGI2026}.
\subsection{ECG-Based Laboratory Value Prediction}
Another area of ECG-based deep learning research investigates whether ECG recordings can be used to predict laboratory values and detect laboratory abnormalities. Serum potassium was among the laboratory values investigated in early ECG-based deep learning research. This is because changes in potassium concentrations can affect cardiac electrophysiology and ECG morphology\cite{arnsdorfELECTROCAR1976, ettingerHYPERKALEM1974}. These effects may include changes in ECG waves and cardiac conduction, although their occurrence and severity vary between different patients\cite{teymouriECGFREQUE2022}.

Subsequently, a deep learning model was developed to detect hyperkalemia from ECG recordings of patients with chronic kidney disease. The model was trained on a large number of ECG recordings and achieved AUC values ranging from 0.853 to 0.883\cite{gallowayDEVELOPMEN2019}. These results indicate that ECG-based deep learning may support noninvasive screening for elevated potassium levels\cite{gallowayDEVELOPMEN2019}.

Recent research has expanded ECG-based prediction to a broad range of laboratory abnormalities\cite{alcaraznormalitypredictionforecasting2025}. A multimodal deep learning approach combined ECG recordings with patient metadata to predict laboratory abnormalities using the MIMIC-IV dataset. Among the investigated biomarkers, NT-proBNP showed the strongest predictive performance, having an AUROC of 0.903\cite{alcaraznormalitypredictionforecasting2025}. Comparatively strong results were also reported for other abnormalities in hemoglobin and albumin\cite{alcaraznormalitypredictionforecasting2025}.

These studies demonstrate that ECG signals may contain information related to a range of laboratory values. However, predictive performance varies between individual biomarkers\cite{alcaraznormalitypredictionforecasting2025}. The positive results for NT-proBNP motivate a more detailed investigation. Although previous work has demonstrated the feasibility of ECG-based laboratory prediction, physiological interpretation and the influence of different technical design choices, such as ECG lead configurations, have received less attention in previous research. Therefore, this work extends previous research by investigating NT-proBNP prediction using regression, binary classification and three-class classification experiments and the effect of 1-lead and 12-lead configurations.
\clearpage
\section{Methodology}
The methodology used in this work was designed to evaluate ECG-based NT-proBNP prediction across different prediction tasks and ECG lead configurations. The following sections describe the methodology used for the experiments. 
\subsection{Problem Formulation} \label{sec:problem-formulation}
This work investigates whether NT-proBNP concentrations can be predicted from electrocardiogram recordings and patient metadata using a multimodal deep learning framework. The main experiments include regression and classification tasks and investigate the influence of using 1-lead and 12-lead settings for both tasks on predictive performance. The classification experiments include classification using a threshold of 300~pg/mL and a three-class task based on the National Institute for Health and Care Excellence (NICE) guidelines\cite{OVERVIEWC2018}.

The regression task aims to estimate the continuous NT-proBNP concentration corresponding to each ECG recording. Regression performance is evaluated using the mean absolute error, Bland--Altman analysis and Passing--Bablok regression.

For the binary classification task, the dataset is divided into two classes. The first class contains NT-proBNP concentrations below 300~pg/mL, whereas the second class contains NT-proBNP concentrations greater than or equal to 300~pg/mL. 

Furthermore, a three-class classification task is defined according to the NICE thresholds\cite{OVERVIEWC2018}. The three classes include NT-proBNP concentrations below 400~pg/mL, concentrations from 400 to 2000~pg/mL inclusive, and concentrations above 2000~pg/mL. Classification performance is evaluated using the area under the receiver operating characteristic curve (AUROC), confusion matrices and additional classification metrics.

The patient metadata is incorporated alongside the ECG recordings as an additional input modality. 

The primary analysis evaluates the regression and classification predictive performance across 1-lead (lead II) and 12-lead ECG configurations. Additional experiments involving hemoglobin, blood glucose, potassium and further NT-proBNP prediction settings are presented in the appendix.

\begin{table}[H]
\centering
\caption{Overview of the primary NT-proBNP experiments.}
\label{tab:experiment_overview}
\begin{tabular}{llll}
\toprule
Task & Thresholds [pg/mL] & Leads & Loss \\
\midrule
Regression & -- & 1-lead  & L1 \\
Regression & -- & 12-lead & L1 \\
Binary classification & 300 & 1-lead  & Weighted focal CE \\
Binary classification & 300 & 12-lead & Weighted focal CE \\
Three-class classification & 400, 2000 & 1-lead  & Weighted focal CE \\
Three-class classification & 400, 2000 & 12-lead & Weighted focal CE \\
\bottomrule
\end{tabular}
\end{table}

Table~\ref{tab:experiment_overview} shows an overview of the primary NT-proBNP experiments and the corresponding prediction tasks, classification thresholds, ECG lead configurations and loss functions.
\subsection{Datasets} \label{sec:datasets}
The data used in this work are obtained from the Medical Information Mart for Intensive Care IV (MIMIC-IV) database and its associated ECG and emergency department datasets. All datasets are obtained through PhysioNet and contain deidentified patient data collected at Beth Israel Deaconess Medical Center in Boston, Massachusetts\cite{PhysioNet-mimiciv-3.1, johnsonMIMICIVFR2023, pollardPHYSIONET2026}.

MIMIC-IV version 3.1 provides the clinical data in this work, including laboratory measurements, patient information and online medical record measurements\cite{PhysioNet-mimiciv-3.1, johnsonMIMICIVFR2023, pollardPHYSIONET2026}. NT-proBNP measurements and the additional laboratory values are obtained from the \texttt{labevents} table. Patient age and sex are extracted from the patients table, while height, weight and body mass index are obtained from the \texttt{omr} table.

The ECG waveforms are obtained from the MIMIC-IV-ECG: Diagnostic Electrocardiogram Matched Subset version 1.0\cite{gowMIMICIVECG, pollardPHYSIONET2026}. The original recordings have a duration of 10 seconds. In this work, segments with a duration of 2.5 seconds are used as model input. ECG records are linked to patients in the MIMIC-IV clinical database, which enables the matching of ECG recordings with laboratory measurements and patient information.

In addition to the waveform data, ECG machine measurements are obtained from MIMIC-IV-ECG. These machine measurements include the RR interval, the onset and end of the P wave and QRS complex, the end of the T wave and the electrical axes of the P wave, QRS complex and T wave.
These machine measurements obtained from MIMIC-IV-ECG are calculated across all 12 leads\cite{gowMIMICIVECG, pollardPHYSIONET2026}.

Vital signs are obtained from MIMIC-IV-ED version 2.2, the database of emergency department admissions at the Beth Israel Deaconess Medical Center between 2011 and 2019\cite{johnsonMIMICIVED, pollardPHYSIONET2026}. MIMIC-IV-ED contains data from emergency department stays and is also linked to patients in the MIMIC-IV clinical database. In this work, the used vital signs include patient temperature, heart rate, respiratory rate, oxygen saturation and systolic and diastolic blood pressure.
\subsection{Data Preprocessing}
The data preprocessing pipeline used in this work combines ECG recordings, laboratory measurements, patient information, ECG machine measurements, online medical record data, and vital signs into a unified dataset for each experiment. Individual data sources are linked using patient and recording identifiers according to the MIMIC-IV documentation\cite{MIMICIV}. Laboratory measurements and ECG recordings are matched based on their temporal proximity. Each resulting sample consists of an ECG recording, the matching target value or classification label, and a feature vector. The dataset is split at the patient level into 90\% training, 5\% validation and 5\% test data to prevent data leakage.
\subsubsection{Laboratory Value and ECG Matching}
NT-proBNP measurements are extracted from the \texttt{labevents} table using the matching item identifier. For laboratory values with multiple item identifiers, one consistent identifier is selected. This was necessary to avoid combining different measurements that may differ in their laboratory procedures or specimen types, such as blood, urine, or other body fluids.

Measurements with missing timestamps, missing, zero or negative numerical target values or missing units are excluded. Measurements without timestamps cannot be accurately matched to an ECG recording, while missing, zero or negative target values do not provide a valid target for either prediction task. Missing target values are not imputed to avoid distorting the target distribution by introducing artificial labels. Furthermore, for the regression experiments, NT-proBNP measurements above 3000~pg/mL are excluded to limit the target range. The remaining regression target values are then standardized using a \texttt{StandardScaler} fitted on the training split. 

\begin{figure}[H]
    \centering
    \includegraphics[width=\textwidth]{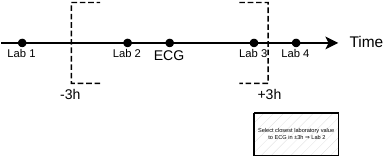}
    \caption{Schematic overview of the temporal matching of ECG and laboratory values.}
    \label{fig:ecg_lab_matching}
\end{figure}

ECG recordings and laboratory measurements from the same patient are matched using a temporal window of $\pm$3 hours, as shown in Figure~\ref{fig:ecg_lab_matching}. This window was chosen to retain a sufficient number of samples while still allowing the most appropriate laboratory measurement to be matched to each ECG recording. For each ECG recording, the laboratory measurement with the smallest absolute temporal distance is selected. If multiple laboratory measurements with the same temporal distance are present, the earlier laboratory timestamp and subsequently the laboratory identifier are used to resolve ties. Duplicate assignments of the same laboratory measurement are removed so individual laboratory values are only represented once in the dataset.

In addition, machine measurements are included in the feature vector. These measurements are described in Section \ref{sec:datasets}. Temporal information is represented by the time difference between the ECG recording and the associated laboratory measurement in minutes and hours. The time of day of the ECG recording is additionally encoded using sine and cosine transformation to preserve its cyclic structure.
\subsubsection{Feature Processing}
Patient age is calculated at the time of the ECG recordings using the anchor age and anchor year of the patient, while the patient sex is extracted and encoded as a binary variable.

Patient height, weight and body mass index (BMI) are obtained from the \texttt{omr} table, while the vital signs are obtained from the \texttt{vitalsign} table. Both data sources are matched to the ECG recording using the same temporal matching procedure as for laboratory measurements. For vital signs, a temporal window of $\pm$1 hour is selected so that the patient condition at the time of the ECG recording is reflected. For body measurements, a temporal window of $\pm$183 days is selected since these measurements are recorded less frequently than laboratory values and vital signs.

Missing values for BMI, height, weight and vital signs are replaced with the corresponding median calculated exclusively from the training split. Median imputation is used to retain more samples with incomplete metadata. Calculating the median only from the training split prevents the validation and test splits from influencing preprocessing. Furthermore, binary missing flags are included so that the model can distinguish between original values and imputed values.

\begin{table}[H]
\centering
\caption{Tabular input features}
\label{tab:tabular_features}
\footnotesize
\setlength{\tabcolsep}{4pt}
\renewcommand{\arraystretch}{0.9}
\begin{tabular}{p{0.31\textwidth} p{0.24\textwidth} p{0.34\textwidth}}
\toprule
Feature & Source & Processing \\
\midrule
Age & MIMIC-IV & Calculated at ECG time \\
Sex & MIMIC-IV & Binary encoding \\
RR interval & MIMIC-IV-ECG & Machine measurement \\
P wave onset & MIMIC-IV-ECG & Machine measurement \\
P wave end & MIMIC-IV-ECG & Machine measurement \\
QRS onset & MIMIC-IV-ECG & Machine measurement \\
QRS end & MIMIC-IV-ECG & Machine measurement \\
T wave end & MIMIC-IV-ECG & Machine measurement \\
P wave axis & MIMIC-IV-ECG & Machine measurement \\
QRS axis & MIMIC-IV-ECG & Machine measurement \\
T wave axis & MIMIC-IV-ECG & Machine measurement \\
BMI & MIMIC-IV & $\pm$183 day matching\\
Height & MIMIC-IV & $\pm$183 day matching\\
Weight & MIMIC-IV & $\pm$183 day matching\\
Time difference [min] & Derived & ECG-laboratory difference \\
Time difference [h] & Derived & ECG-laboratory difference \\
ECG time sine & Derived & Cyclic encoding \\
ECG time cosine & Derived & Cyclic encoding \\
BMI missing & Derived & Binary missing indicator \\
Height missing & Derived & Binary missing indicator \\
Weight missing & Derived & Binary missing indicator \\
Temperature & MIMIC-IV-ED & $\pm$1 hour matching \\
Heart rate & MIMIC-IV-ED & $\pm$1 hour matching \\
Respiratory rate & MIMIC-IV-ED & $\pm$1 hour matching \\
Oxygen saturation & MIMIC-IV-ED & $\pm$1 hour matching \\
Systolic blood pressure & MIMIC-IV-ED & $\pm$1 hour matching \\
Diastolic blood pressure & MIMIC-IV-ED & $\pm$1 hour matching \\
Temperature missing & Derived & Binary missing indicator \\
Heart rate missing & Derived & Binary missing indicator \\
Respiratory rate missing & Derived & Binary missing indicator \\
Oxygen saturation missing & Derived & Binary missing indicator \\
Systolic BP missing & Derived & Binary missing indicator \\
Diastolic BP missing & Derived & Binary missing indicator \\
\bottomrule
\end{tabular}
\end{table}

Table~\ref{tab:tabular_features} summarizes the 33 features used in the experiments.
\subsection{Model Architecture}
The model architecture used in this work is based on the MDS-ICU framework developed by López Alcaraz et al. \cite{alcarazMULTIMODAL2026}. The framework was provided for this thesis and adapted for the prediction of different laboratory values. It consists of separate encoder branches for ECG waveforms and tabular features. The ECG branch consists of four stacked structured state space (S4) blocks followed by a pooling head, while the tabular branch uses a RealMLP encoder. The representations generated by both branches are then combined using the late fusion described in MDS-ICU\cite{alcarazMULTIMODAL2026}.

In this work, the ECG recordings were resampled to a sampling frequency of 100~Hz and divided into segments of 2.5 seconds, resulting in 250 samples per lead. The 1-lead configuration uses lead II as the waveform input, while the 12-lead configuration uses all available ECG leads. The tabular branch receives the 33 features derived during data preprocessing. These features include demographic information, machine-generated ECG measurements, body measurements, vital signs, temporal features and missing flags for the imputed body measurements and vital signs. The underlying encoder and fusion architecture remain unchanged across these experiments.
\newline

\begin{figure}[H]
    \centering
    \includegraphics[width=\textwidth]{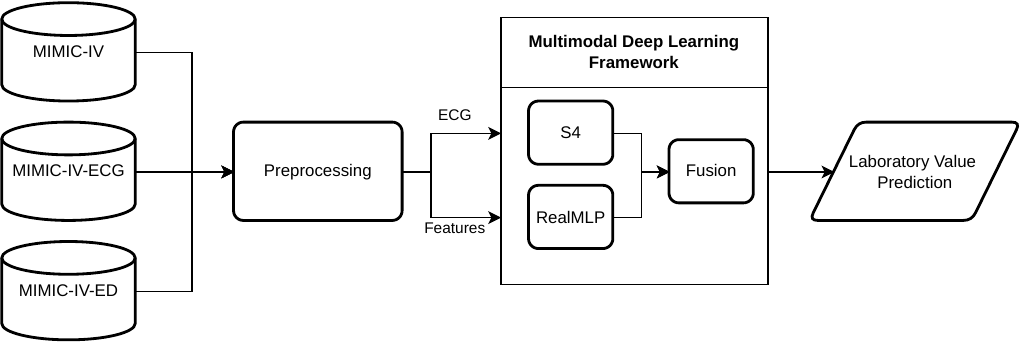}
    \caption{Overview of the pipeline for the laboratory value prediction.}
    \label{fig:methodology_overview}
\end{figure}

Figure~\ref{fig:methodology_overview} shows an overview of the complete pipeline from the datasets to laboratory value prediction.
\subsection{Training Procedure}
The models in this work are trained from scratch. The general optimization settings remain unchanged for the regression and classification experiments and the 1-lead and 12-lead configurations. The complete training configuration is shown in Table~\ref{tab:training_configuration}.

L1 loss is used to optimize the regression models based on the difference between the predicted and standardized target values. Parameters used for target standardization are calculated only from the training split. For evaluation, the predictions are transformed back into the original unit of measurement of pg/mL. 

For the classification experiments, focal cross entropy loss is used to reduce the contribution of easily classified samples and focus more on difficult examples. Furthermore, class weights are added to account for the imbalance in the training data.

\begin{table}[H]
\centering
\caption{Training configuration for the primary NT-proBNP experiments.}
\label{tab:training_configuration}
\begin{tabular}{ll}
\toprule
Parameter & Value \\
\midrule
Epochs & 20 \\
Batch size & 32 \\
Optimizer & AdamW \\
Learning rate & 0.001 \\
Weight decay & 0.001 \\
Precision & 32-bit \\
Regression loss & L1 loss \\
Classification loss & Focal cross entropy loss \\
Focusing parameter & 2 \\
Binary class weights & 2.2414, 0.6436 \\
Three-class weights & 1.2811, 1.1668, 0.7340 \\
\bottomrule
\end{tabular}
\end{table}
\subsection{Evaluation Metrics}
Different evaluation metrics are used depending on the prediction task. Regression performance is evaluated using the mean absolute error, Bland--Altman analysis and Passing--Bablok regression. Classification performance is evaluated using the area under the receiver operating characteristic curve (AUROC), accuracy, sensitivity, F1 score and confusion matrices. In addition, 95\% confidence intervals are used to indicate the precision of the reported performance estimates. Following the evaluation procedure used in the MDS-ICU framework\cite{alcarazMULTIMODAL2026}, confidence intervals are estimated using 1000 bootstrap iterations.
\subsubsection{Regression Evaluation} \label{sec:regression_evaluation}
Mean absolute error (MAE) measures the average absolute difference between the predicted and true NT-proBNP concentrations and is used in this work as the primary performance metric for the regression experiments. It is defined as

\begin{equation}
    \mathrm{MAE}
    =
    \frac{1}{N}
    \sum_{i=1}^{N}
    \left| y_i - \hat{y}_i \right|,
\end{equation}

where \(N\) denotes the number of samples, \(y_i\) the true NT-proBNP concentration, and \(\hat{y}_i\) the corresponding model prediction. Lower MAE values indicate smaller prediction errors. Since MAE is reported in the same unit as the target, it allows the prediction error to be interpreted directly in the original NT-proBNP scale.

Due to the standardization of the regression targets, the MAE reported by the training framework is initially calculated on the standardized targets. For the final evaluation, the predictions are transformed back to the original scale, and the MAE is reported in pg/mL.

In addition to MAE, Bland--Altman analysis, using the \texttt{statsmodels} Python package \cite{seabold2010statsmodels}, is used to analyze how the predicted NT-proBNP concentration compares to the matching target values. For each sample, the difference between the predicted concentration and reference value is plotted against the mean of both values. The mean difference indicates whether the model tends to overestimate or underestimate NT-proBNP concentrations. 

The limits of agreement are calculated as

\begin{equation}
    \bar{d} \pm 1.96 s_d,
\end{equation}

where \(\bar{d}\) denotes the mean difference between predicted and reference values and \(s_d\) is the standard deviation of these differences. The limits of agreement show the range in which approximately 95\% of the differences between predicted and reference values are expected to lie. Bland--Altman analysis complements MAE by providing additional information about the deviation and spread of differences between the predicted NT-proBNP concentration and the reference concentration.

For further evaluation of the relationship between predicted and reference NT-proBNP concentrations, Passing--Bablok regression is used. This method is used to estimate a regression line between both measurements and show systematic and proportional differences and therefore complements the previous analyses. For the analysis, the \texttt{mcr} package in R is used with the bootstrap quantile method and 1000 bootstrap samples. Since the Passing--Bablok implementation in \texttt{mcr} does not support negative values, the negative model predictions are excluded from this analysis.
\subsubsection{Classification Evaluation}
The primary evaluation metric for the classification experiments is the area under the receiver operating characteristic curve (AUROC). For the three-class classification task, macro-averaged AUROC is used. The receiver operating characteristic curve demonstrates how the true positive and false positive rates change across different classification thresholds, while AUROC summarizes this relationship into a single value. Higher AUROC values indicate better classification performance, with a value of 0.5 indicating no discriminative ability and a value of 1 indicating perfect discrimination. Additionally, accuracy, sensitivity and F1 score are reported and calculated using the \texttt{scikit-learn} library.

Sensitivity and F1 score are calculated in the binary classification evaluation for the positive class of NT-proBNP concentrations greater than or equal to 300~pg/mL. Furthermore, the three-class classification evaluation uses macro-averaged sensitivity and F1 score to give every class equal weighting regardless of class size. 

Furthermore, confusion matrices, using the \texttt{scikit-learn} library\cite{scikit-learn}, are used to evaluate the distribution of correct and incorrect class predictions, where the predicted class labels are compared to the corresponding reference labels to show how frequently samples from each class are assigned to the different classes. This complements AUROC by showing the distribution of classification errors.
\subsection{Additional Laboratory Value Experiments}
In addition to the primary NT-proBNP experiments, additional experiments for hemoglobin, glucose and potassium were conducted. These experiments investigated whether the multimodal ECG-based prediction approach could also be applied to other laboratory values. Furthermore, alternative classification and regression configurations for NT-proBNP were investigated. The detailed experimental setups and results are presented in Appendix~\ref{sec:additional_ntprobnp}, Appendix~\ref{sec:hemoglobin}, Appendix~\ref{sec:glucose} and Appendix~\ref{sec:potassium}.
\clearpage
\section{Experiments and Results}
This section presents the results of the regression and classification experiments for the prediction of NT-proBNP from ECG signals and patient metadata. The experiments are evaluated using the metrics described in the previous section. In addition, the predictive performance of 1-lead and 12-lead ECG configurations is compared.

\subsection{Data Description}
\begin{table}[H]
\centering
\caption{Test split size for the primary NT-proBNP experiments.}
\label{tab:dataset_size}
\begin{tabular}{lcc}
\toprule
Task & ECG recordings & ECG segments \\
\midrule
Regression     & 910  & 3640 \\
Classification & 1435 & 5740 \\
\bottomrule
\end{tabular}
\end{table}

The final datasets differed between the regression and classification experiments, since NT-proBNP concentrations above 3000~pg/mL were excluded from the regression dataset. Table~\ref{tab:dataset_size} shows the number of ECG recordings and ECG segments in the test splits for each task. Each ECG recording was divided into four 2.5 second segments and all reported prediction metrics were calculated using the 2.5 second ECG segments.
\begin{table}[H]
\centering
\caption{Test split class distribution for the NT-proBNP classification experiments.}
\label{tab:class_distribution}
\begin{tabular}{lccc}
\toprule
Task & Class & ECG recordings & ECG segments \\
\midrule
Binary & $<300$~pg/mL & 307 & 1228 \\
  & $\geq300$~pg/mL & 1128 & 4512 \\
\midrule
Three-class & $<400$~pg/mL & 381 & 1524 \\
  & $400$--$2000$~pg/mL & 410 & 1640 \\
  & $>2000$~pg/mL & 644 & 2576 \\
\bottomrule
\end{tabular}
\end{table}

Table~\ref{tab:class_distribution} shows the class distribution in the test splits for the binary and three-class classification experiments. The binary classification task showed a stronger class imbalance than the three-class classification task.
\begin{table}[H]
\centering
\caption{Missing metadata counts in the test splits for the primary NT-proBNP experiments.}
\label{tab:missing_metadata}
\begin{tabular}{lcc}
\toprule
Feature & Regression n (\%) & Classification n (\%) \\
\midrule
BMI                     & 229 (25.2\%)  & 418 (29.1\%) \\
Height                  & 367 (40.3\%)  & 653 (45.5\%) \\
Weight                  & 187 (20.5\%)  & 325 (22.6\%) \\
Temperature             & 628 (69.0\%)  & 1057 (73.7\%) \\
Heart rate              & 470 (51.6\%)  & 776 (54.1\%) \\
Respiratory rate        & 472 (51.9\%)  & 781 (54.4\%) \\
Oxygen saturation       & 487 (53.5\%)  & 806 (56.2\%) \\
Systolic blood pressure & 480 (52.7\%)  & 787 (54.8\%) \\
Diastolic blood pressure& 480 (52.7\%)  & 787 (54.8\%) \\
\bottomrule
\end{tabular}
\end{table}

Table~\ref{tab:missing_metadata} shows the number of missing metadata values in the test splits for both tasks. These missing values were imputed using the median calculated from the training splits.
\subsection{NT-proBNP Regression}
The following results focus on the regression experiments and compare the 1-lead and 12-lead ECG configurations regarding prediction error and agreement with the reference NT-proBNP concentrations.
\newline

\begin{table}[H]
    \centering
    \caption{Mean absolute error with 95\% confidence intervals for NT-proBNP prediction.}
    \label{tab:ntprobnp_regression}
    \begin{tabular}{lcc}
        \hline
        ECG configuration & MAE [pg/mL] & 95\% CI [pg/mL] \\
        \hline
        1-lead  & 534.98 & 501.16--571.62 \\
        12-lead & 514.27 & 482.83--546.46 \\
        \hline
    \end{tabular}
\end{table}

Table~\ref{tab:ntprobnp_regression} shows the regression performance of the 1-lead and 12-lead experiments. The 1-lead model achieved an MAE of 534.98~pg/mL (95\% CI: 501.16--571.62~pg/mL), while the 12-lead model achieved an MAE of 514.27~pg/mL (95\% CI: 482.83--546.46~pg/mL). The MAE of the 12-lead configuration was therefore 20.71~pg/mL lower than that of the 1-lead configuration.

\begin{figure}[H]
    \centering

    \begin{subfigure}[t]{0.65\textwidth}
        \centering
        \includegraphics[width=\textwidth]{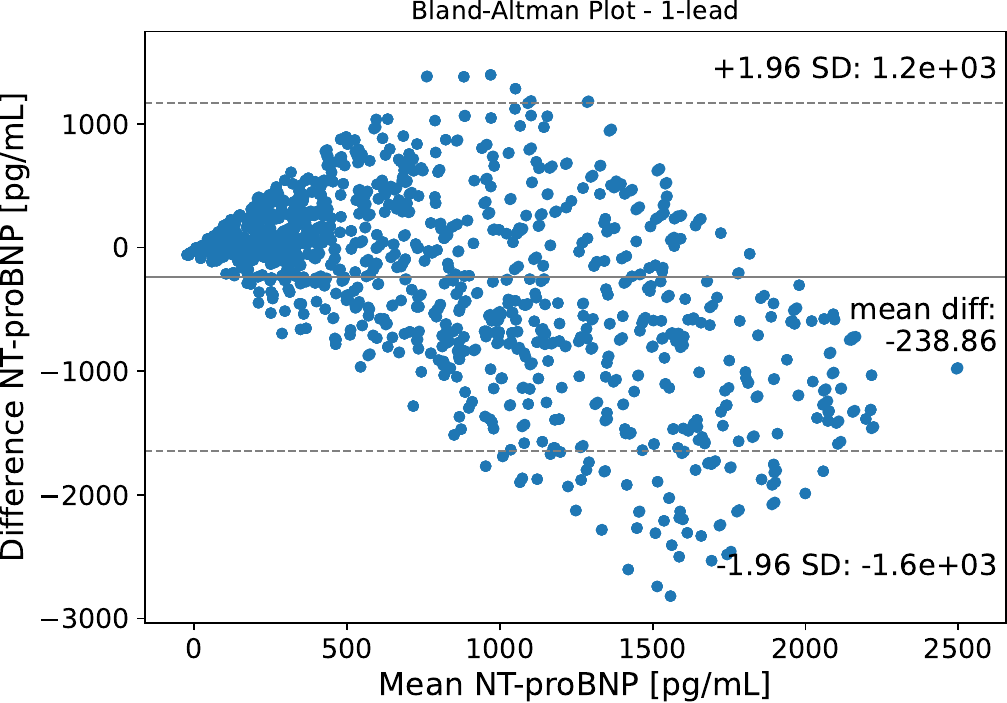}
        \caption{1-lead ECG}
        \label{fig:ba_1lead}
    \end{subfigure}
    \hfill
    \begin{subfigure}[t]{0.65\textwidth}
        \centering
        \includegraphics[width=\textwidth]{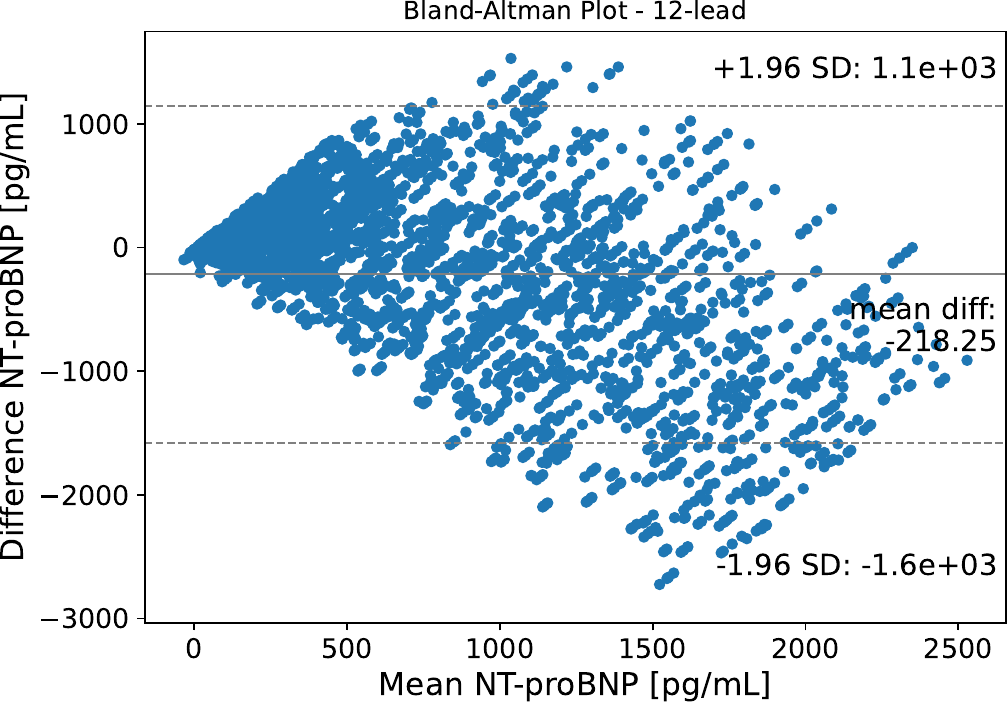}
        \caption{12-lead ECG}
        \label{fig:ba_12lead}
    \end{subfigure}

    \caption{Bland--Altman analysis of predicted and reference NT-proBNP concentrations for the 1-lead and 12-lead configurations. The solid line represents the mean difference and the dashed lines indicate the 95\% limits of agreement.}
    \label{fig:bland_altman}
\end{figure}

Figure~\ref{fig:bland_altman} illustrates the Bland--Altman analysis for both ECG configurations. The 1-lead model showed a mean difference of -238.86~pg/mL with limits of agreement from -1646.41 to 1168.69~pg/mL, while the 12-lead model showed a mean difference of -218.25~pg/mL with limits of agreement from -1583.4 to 1146.9~pg/mL. In both configurations, the spread of the differences widened at higher NT-proBNP concentrations.

\begin{figure}[H]
\centering

\begin{subfigure}[t]{0.65\textwidth}
    \centering
    \includegraphics[width=\textwidth]{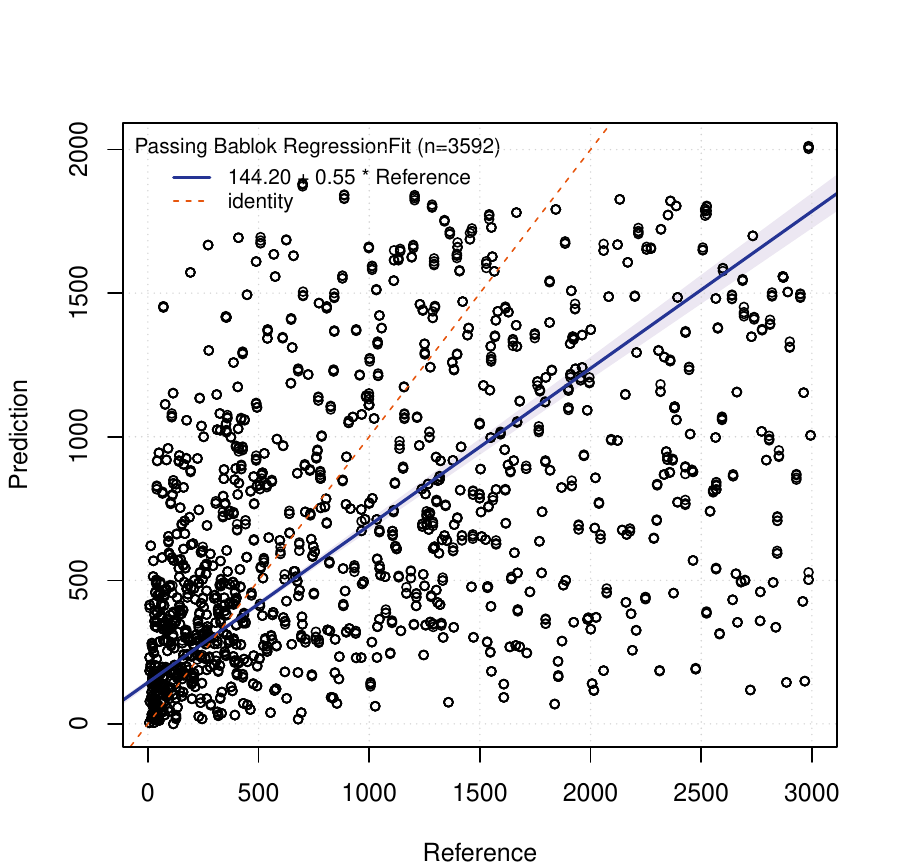}
    \caption{1-lead ECG}
    \label{fig:pb_1lead}
\end{subfigure}
\hfill
\begin{subfigure}[t]{0.65\textwidth}
    \centering
    \includegraphics[width=\textwidth]{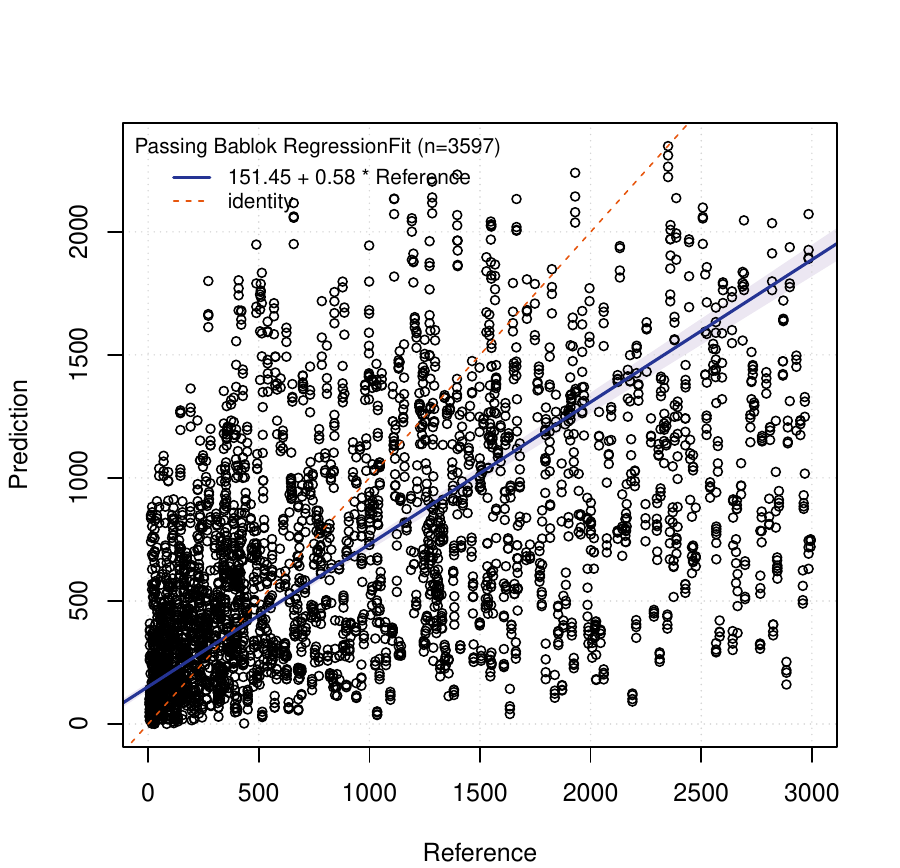}
    \caption{12-lead ECG}
    \label{fig:pb_12lead}
\end{subfigure}

\caption{Passing--Bablok regression of predicted and reference NT-proBNP concentrations for the 1-lead and 12-lead configurations. The solid line represents the Passing--Bablok regression fit, the dashed line represents the identity, and the shaded area represents the 95\% confidence interval.}
\label{fig:passing_bablok}
\end{figure}

Figure~\ref{fig:passing_bablok} shows the Passing--Bablok regression analysis for 1-lead and 12-lead ECG configurations. The intercept for the 1-lead configuration was 144.20~pg/mL (95\% CI: 132.35--153.81~pg/mL) and the slope was 0.55 (95\% CI: 0.53--0.57), while for the 12-lead configuration the intercept was 151.45~pg/mL (95\% CI: 136.64--167.10~pg/mL) and the slope was 0.58 (95\% CI: 0.56--0.60).
\subsection{NT-proBNP Classification}
The following results show the predictive performance of the classification experiments, starting with the binary classification experiment followed by the three-class classification experiment.

\subsubsection{Binary Classification}

\begin{table}[H]
\centering
\caption{Binary classification performance for NT-proBNP prediction.}
\label{tab:ntprobnp_binary}
\begin{tabular}{lccccc}
\hline
ECG configuration & AUROC & 95\% CI & Accuracy & Sensitivity & F1 score \\
\hline
1-lead  & 0.849 & 0.839--0.860 & 0.770 & 0.784 & 0.843 \\
12-lead & 0.876 & 0.866--0.886 & 0.799 & 0.798 & 0.862 \\
\hline
\end{tabular}
\end{table}

Table~\ref{tab:ntprobnp_binary} summarizes the performance of the binary classification experiments. The 1-lead experiment achieved an AUROC of 0.849 (95\% CI: 0.839--0.860), while the 12-lead experiment achieved an AUROC of 0.876 (95\% CI: 0.866--0.886). The 12-lead configuration also showed higher values across all other metrics.

\begin{figure}[H]
\centering

\begin{subfigure}[t]{0.65\textwidth}
    \centering
    \includegraphics[width=\textwidth]{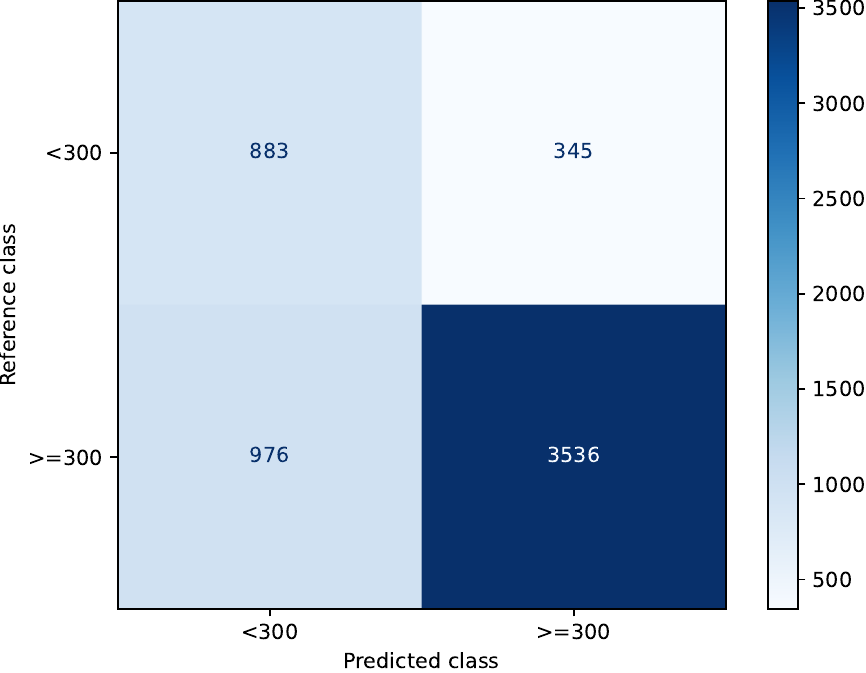}
    \caption{1-lead ECG}
    \label{fig:cm_binary_1lead}
\end{subfigure}
\hfill
\begin{subfigure}[t]{0.65\textwidth}
    \centering
    \includegraphics[width=\textwidth]{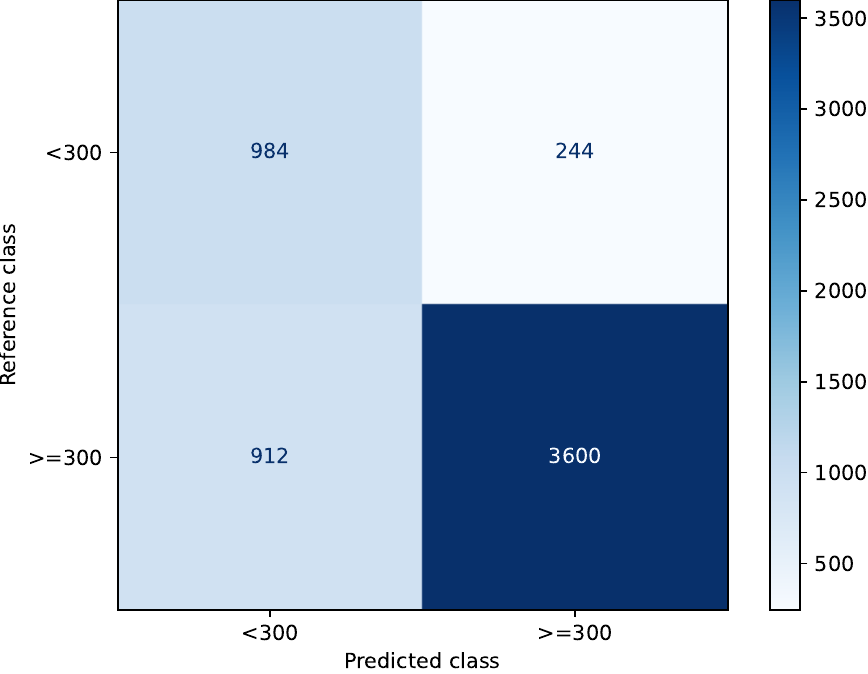}
    \caption{12-lead ECG}
    \label{fig:cm_binary_12lead}
\end{subfigure}

\caption{Confusion matrices for the binary NT-proBNP classification task using the 1-lead and 12-lead ECG configurations.}
\label{fig:cm_binary}
\end{figure}

Figure~\ref{fig:cm_binary} shows the confusion matrices for both ECG configurations. The 1-lead configuration correctly classified 883 samples below 300~pg/mL and 3536 samples greater than or equal to 300~pg/mL. The 12-lead configuration correctly classified 984 samples below 300~pg/mL and 3600 samples greater than or equal to 300~pg/mL. The number of false positive classifications decreased from 345 to 244, while the false negative classifications decreased from 976 to 912.
\subsubsection{Three-Class Classification} \label{sec:three-class}
\begin{table}[H]
\centering
\caption{Three-class classification performance for NT-proBNP prediction.}
\label{tab:ntprobnp_threeclass}
\begin{tabular}{lccccc}
\hline
ECG configuration & AUROC & 95\% CI & Accuracy & Sensitivity & F1 score \\
\hline
1-lead  & 0.738 & 0.729--0.747 & 0.551 & 0.527 & 0.524 \\
12-lead & 0.752 & 0.743--0.761 & 0.578 & 0.565 & 0.539  \\
\hline
\end{tabular}
\end{table}

Table~\ref{tab:ntprobnp_threeclass} summarizes the performance of the three-class classification experiment, based on the NICE guidelines. The 1-lead configuration achieved a macro-averaged AUROC of 0.738 (95\% CI: 0.729--0.747) and the 12-lead configuration achieved a macro-averaged AUROC of 0.752 (95\% CI: 0.743--0.761). The 12-lead configuration also achieved higher accuracy, macro-averaged sensitivity and macro-averaged F1 score.

\begin{figure}[H]
\centering

\begin{subfigure}[t]{0.65\textwidth}
    \centering
    \includegraphics[width=\textwidth]{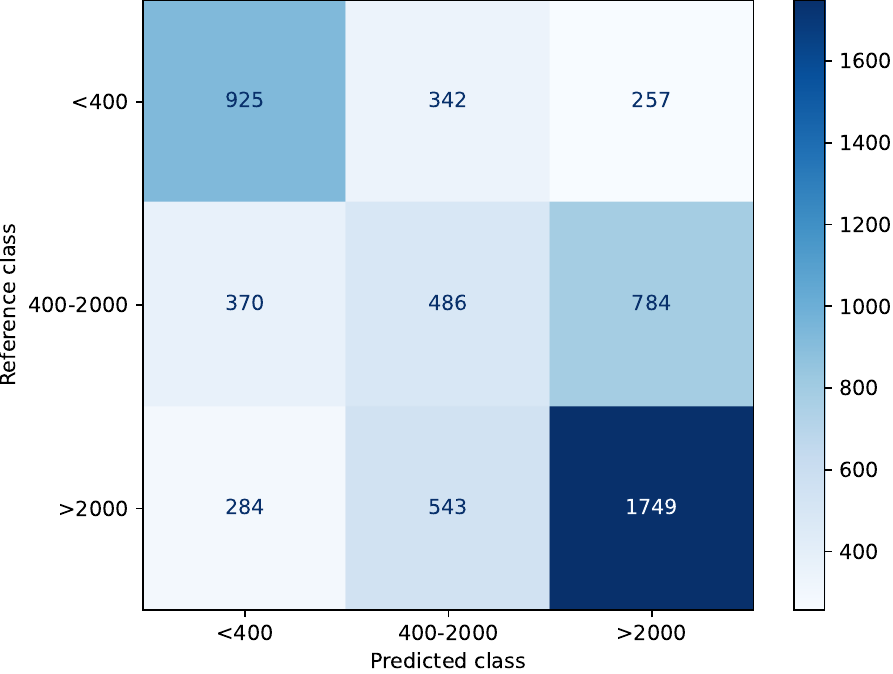}
    \caption{1-lead ECG}
    \label{fig:cm_threeclass_1lead}
\end{subfigure}
\hfill
\begin{subfigure}[t]{0.65\textwidth}
    \centering
    \includegraphics[width=\textwidth]{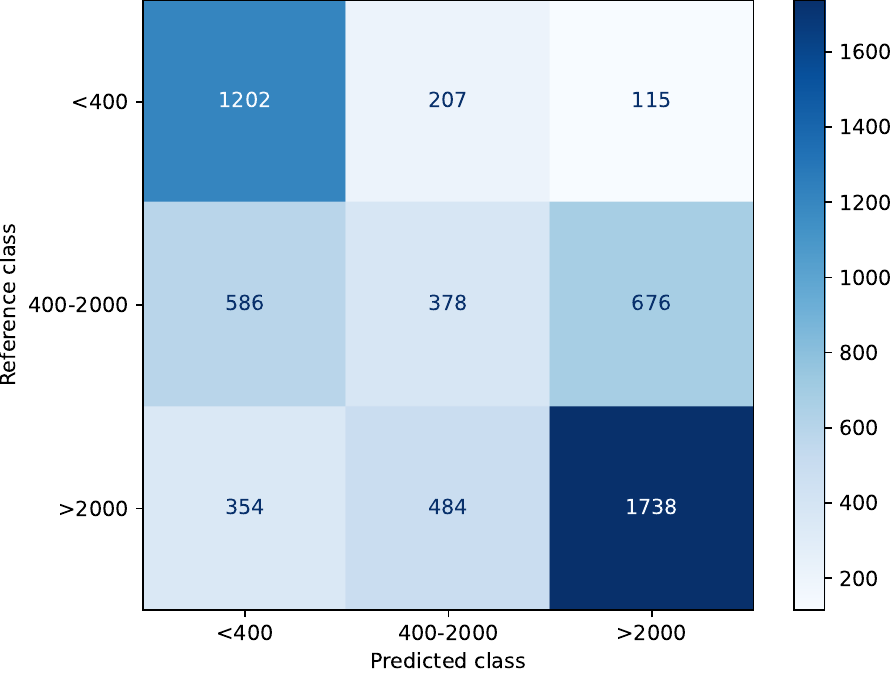}
    \caption{12-lead ECG}
    \label{fig:cm_threeclass_12lead}
\end{subfigure}

\caption{Confusion matrices for the three-class NT-proBNP classification task using the 1-lead and 12-lead ECG configurations.}
\label{fig:cm_threeclass}
\end{figure}

Figure~\ref{fig:cm_threeclass} illustrates the confusion matrices for the three-class NT-proBNP classification task, based on the NICE guidelines. The 1-lead configuration correctly classified 925 samples below 400~pg/mL, 486 between 400 and 2000~pg/mL and 1749 samples above 2000~pg/mL. The 12-lead configuration correctly classified 1202, 378 and 1738 respectively. The 400--2000~pg/mL class showed the lowest proportion of correct classifications for both ECG configurations.
\subsection{Additional Laboratory Value Results}
The additional experiments showed varying predictive performance across the investigated laboratory values. Hemoglobin achieved mean absolute errors of approximately 1.60~g/dL for both ECG configurations. Glucose achieved a mean absolute error of 34.52~mg/dL and a macro-averaged AUROC of 0.632 for the three-class classification experiment. Potassium achieved a mean absolute error of 0.51~mEq/L for the regression experiment and a macro-averaged AUROC of 0.629 for the classification experiment.

The additional NT-proBNP experiments compared alternative classification and regression configurations. The four-class classification experiment showed that weighted cross entropy achieved the highest macro-averaged AUROC of 0.726, compared with 0.714 for unweighted cross entropy and 0.709 for weighted focal cross entropy. Standard scaling achieved a lower MAE than MinMax scaling in the regression experiments, with 2317.63~pg/mL compared with 2390.46~pg/mL.

Detailed results of these experiments are presented in Appendix~\ref{sec:additional_ntprobnp}, Appendix~\ref{sec:hemoglobin}, Appendix~\ref{sec:glucose} and Appendix~\ref{sec:potassium}.
\subsection{Summary}
Across all evaluation tasks, the 12-lead ECG configuration consistently achieved better predictive performance than the 1-lead configuration, although the improvements remained moderate. The three-class classification task showed lower predictive performance compared to the binary classification task. The 400--2000~pg/mL class showed the lowest proportion of correctly classified samples for both ECG lead configurations.
\clearpage
\section{Discussion}
In this chapter, the experimental results are discussed from physiological, technical and clinical perspectives. Furthermore, the results are compared with previous research, the limitations of this work are discussed and possible future research directions are documented.
\subsection{Physiological Interpretation} \label{sec:physiological_interpretation}
The results of this work suggest that information associated with NT-proBNP concentration may be extracted from ECG signals in combination with patient metadata. From a physiological point of view, this relationship is plausible because NT-proBNP is closely associated with cardiac stress and dysfunction\cite{nishikimiBTYPENATR2022}. However, NT-proBNP should not be viewed as directly causing changes in the ECG, but elevated concentrations of NT-proBNP and altered ECG characteristics may reflect shared underlying cardiovascular processes.

Increased myocardial wall stress, which is often associated with elevated NT-proBNP concentrations, can occur as a consequence of cardiac pressure and may be accompanied by functional or structural changes of the heart\cite{nishikimiBTYPENATR2022}. These changes can include alterations in ventricular and atrial dimensions\cite{teramotoLONGITUDIN2024}. Since such alterations can influence the electrical activity of the heart, they provide a possible physiological link between ECG changes and elevated NT-proBNP concentrations.

Furthermore, alterations of myocardial structure and ventricular function can affect the propagation and recovery of cardiac electrical activity, which may result in changes in cardiac conduction and repolarization\cite{magesELECTRICAL2021}. These changes may then become visible as changes in the ECG waveform\cite{hombachELECTROCAR2006}. This further explains why cardiovascular conditions associated with elevated NT-proBNP concentrations may leave patterns in ECG recordings.

This basis is also consistent with the classification results documented in this work. The ability of the model to distinguish between lower and elevated NT-proBNP concentrations suggests that cardiovascular states associated with increased NT-proBNP concentrations may be linked to detectable patterns in the ECG signals and patient metadata. These patterns may reflect underlying changes in cardiac physiology instead of NT-proBNP concentration itself. The classification results therefore support the physiological plausibility that elevated NT-proBNP and altered cardiovascular characteristics may be related through shared underlying physiological processes.

The lower performance of the three-class classification experiments may also be explained by the physiological relationship. NT-proBNP is measured on a continuous scale and is associated with various degrees of cardiac stress and dysfunction, whereas the NICE thresholds divide this continuous concentration into discrete groups\cite{OVERVIEWC2018}. Patients with NT-proBNP values close to a class boundary may therefore be assigned to different classes despite showing similar cardiovascular characteristics. This similarity in physiological characteristics between patients from neighboring classification ranges may then complicate the classification between the three NT-proBNP classes.

In addition, this similarity in physiological characteristics may be particularly relevant for the 400--2000~pg/mL class, which showed the lowest proportion of correctly classified samples in both ECG configurations. Patients in this range may differ considerably in their cardiovascular condition despite showing similar NT-proBNP concentrations. This physiological heterogeneity could have contributed to the lower classification performance for the 400--2000~pg/mL class.

Unlike the classification experiments, the regression experiments show a different physiological challenge. The model is required to predict the exact NT-proBNP concentration from the available ECG signals and patient information. Accurate regression requires the available physiological information to be sufficiently informative about the measured biomarker concentration. However, NT-proBNP concentrations are influenced by multiple physiological and patient-specific factors\cite{bayes-genisPRACTICAL2023}, which may limit the precision of the concentration prediction.

At higher NT-proBNP concentrations, the regression error showed a wider spread, indicating less precise predictions in this range. One possible physiological explanation is that high NT-proBNP concentrations may not correspond to one specific physiological or electrocardiographic pattern. Variability in the cardiovascular conditions associated with elevated NT-proBNP may therefore contribute to the difficulty of predicting precise biomarker concentrations from ECG signals.

The relationship between NT-proBNP concentration and cardiac physiology is also influenced by patient-specific factors\cite{bayes-genisPRACTICAL2023}. Some of these factors are represented in the patient metadata used by the model and may therefore have provided additional context for the prediction of NT-proBNP concentration. 

To summarize, the findings are physiologically consistent with the hypothesis that ECG signals and patient metadata contain indirect information associated with NT-proBNP elevation. However, since NT-proBNP concentrations are influenced by multiple physiological and patient-specific factors, the prediction of precise concentration values remains challenging.
\subsection{Technical Interpretation}
The results in this work show a consistent improvement in the predictive performance of the 12-lead ECG configurations compared to the 1-lead configurations in all evaluated prediction tasks. Specifically, the 12-lead model achieved lower prediction error in the regression experiments and higher AUROC values in the classification experiments. The 1-lead configuration in this work used lead II, which only represented the electrical activity from one spatial perspective\cite{kligfieldRECOMMENDA2007}. The additional spatial information may have been one reason for the improvement in the predictive performance. Additionally, ECG characteristics that may be less apparent in lead II may be more clearly visible in the 12-lead configuration.

Despite the consistent improvement of the 12-lead configuration, the actual improvement remained relatively moderate across the evaluated tasks. The regression experiment showed a mean absolute error decrease from 534.98~pg/mL to 514.27~pg/mL, while the AUROC increased in the binary classification experiment from 0.849 to 0.876 and in the three-class classification experiment from 0.738 to 0.752. These results suggest that the 1-lead multimodal configuration already provided a considerable amount of information relevant to the prediction task. At the same time, this improvement comes with a more complex ECG waveform configuration.

Despite the difference in regression performance, the experiments showed a similar error pattern for both models. The Bland--Altman analysis showed a negative mean difference for the 1-lead and 12-lead configurations, indicating an overall tendency for the model to predict lower NT-proBNP concentrations than the corresponding target values. Both configurations also showed a similar spread of prediction errors. Since the general error pattern was documented for both ECG configurations, using the 12-lead compared to using the 1-lead did not substantially change the general regression behavior.

The Passing--Bablok regression further supports this claim. Both ECG configurations showed slopes clearly below one and positive intercepts, indicating that the predicted NT-proBNP concentrations increased more slowly than the associated laboratory values. While the 12-lead configuration showed a slightly higher intercept, its slope was slightly closer to one. Overall, both configurations demonstrated a similar pattern of deviation from the target values, indicating further that using additional leads may not change the regression error pattern substantially.

The binary classification results showed that the improvement in predictive performance by using the 12-lead configuration was not limited to a single evaluation metric. In addition to improvements in AUROC, accuracy, sensitivity and F1 score also improved slightly. These improvements across different evaluation metrics suggest that the 12-lead configuration may support a more consistent separation between lower and elevated NT-proBNP concentrations.

The three-class classification task also showed improvements, although they were less evident. The 12-lead configuration showed a higher AUROC, accuracy, sensitivity and F1 score. However, the improvement was not equally distributed across all NT-proBNP classes. More samples were correctly classified in the below 400~pg/mL class, while the 400--2000~pg/mL and above 2000~pg/mL classes showed a decrease in correctly classified samples. This indicates that an improvement in overall classification performance does not necessarily result in more correctly classified samples for every class. 

Overall, the results show that the model achieved better predictive performance across all experiments using the 12-lead ECG configuration compared to using the 1-lead configuration. However, this improvement is only moderate. Furthermore, both regression experiments showed similar predictive deviations from the laboratory values, indicating that the use of the 12-lead configurations compared to the 1-lead configurations did not change the difficulty for the model to predict a precise NT-proBNP concentration.
\subsection{Comparison with Previous Work}
The results of this work are broadly consistent with previous research indicating that ECG recordings with patient metadata contain information related to laboratory value prediction. López Alcaraz and Strodthoff investigated the prediction of multiple laboratory abnormalities from ECG signals and clinical data, also using the MIMIC-IV database. Among the investigated laboratory values, NT-proBNP achieved the highest predictive performance, with an AUROC of 0.903 for the prediction of concentrations above 353~pg/mL\cite{alcaraznormalitypredictionforecasting2025}. The binary classification experiments in this work also showed that elevated NT-proBNP concentrations may be distinguished from lower concentrations with an AUROC of 0.876 using the 12-lead configuration.

These performance values should not be compared directly, however, since the experimental settings differ. López Alcaraz and Strodthoff used a threshold of 353~pg/mL, while this work used a threshold of 300~pg/mL for the binary classification task and different dataset construction and temporal matching. However, both works showed that ECG signals combined with patient metadata contain predictive information about elevated NT-proBNP concentrations.

In addition, this work extended the investigation of NT-proBNP to additional prediction tasks, such as regression and three-class classification based on clinically established NICE thresholds\cite{OVERVIEWC2018}. The regression experiments showed that the prediction of precise NT-proBNP concentrations remains challenging, while the three-class classification experiments were able to distinguish between the concentration ranges, albeit the predictive performance was lower than in the binary classification task.

Furthermore, this work extended the prediction tasks with a comparison between 1-lead and 12-lead configurations. While López Alcaraz and Strodthoff used 12-lead ECG recordings, this work additionally evaluated 1-lead configurations with lead II and found consistently better predictive performance with the 12-lead configurations.
\subsection{Clinical Relevance}
The results of this work suggest that ECG-based prediction of NT-proBNP may have potential clinical relevance as an additional source of information about the cardiovascular state of a patient. Acquisition of ECG is noninvasive, as opposed to the common approach of using invasive blood sampling\cite{lima-oliveiraLABORATORY2015}. Therefore, a trained model could be used with ECG signals and patient metadata to provide additional information to clinicians before laboratory results are available. This information should therefore be considered complementary to a direct laboratory measurement rather than as a replacement for it.

The results of the classification tasks may be particularly relevant from a clinical perspective. The classification tasks used in this work defined concentration groups based on established NT-proBNP thresholds\cite{OVERVIEWC2018}. The model could therefore potentially provide an early indication of whether the NT-proBNP concentration falls within an elevated concentration range before the actual laboratory results are available.

The comparison between the 1-lead configuration and 12-lead configuration is also relevant when considering potential clinical applications. Although consistently better prediction performance was documented in the 12-lead configuration, the difference compared to 1-lead remained moderate, indicating that the 1-lead multimodal configuration already retained considerable predictive performance, which may partly be explained by the ECG machine measurements being calculated using all 12 leads. However, if a 12-lead ECG is already available, the results suggest that using the 12-lead waveform input may lead to better predictive performance compared to 1-lead.

Another potential clinical use for this model may be as a preliminary indicator for elevated NT-proBNP concentrations. The information from the model could then be used as an additional indicator of whether further clinical assessment may be warranted for the patient. While the regression results showed limited predictive performance for precise NT-proBNP concentration prediction, the classification experiments showed that the model could distinguish between concentration groups. This indicates that the model may be used as additional decision support rather than a diagnostic measurement. 

Multimodal approaches combining ECG recordings with clinical information have been investigated for clinical decision support in intensive care units and emergency departments\cite{alcarazMULTIMODAL2026, alcarazhancingclinicaldecision2025}. These settings may be particularly relevant since ECG recordings and other patient information are routinely available. However, further validation is required before the NT-proBNP prediction approach investigated in this work could be applied in such clinical settings.

Overall, ECG-based NT-proBNP prediction may be most useful as a complementary source of information. In particular, the classification of broader concentration ranges may provide useful preliminary information, while the results from the regression experiments indicate that precise prediction of NT-proBNP concentration remains challenging.
\subsection{Generalization to Other Laboratory Values}
The additional experiments on hemoglobin, glucose and potassium showed that the multimodal ECG-based approach can be applied to different laboratory values with varying predictive performance. This is consistent with previous research showing that the predictive performance of ECG-based laboratory value prediction differs between individual biomarkers\cite{alcaraznormalitypredictionforecasting2025}.

The regression experiments showed a similar pattern across several laboratory values. For hemoglobin, glucose and potassium the predicted concentrations varied less than the corresponding reference concentrations. This indicates that the models had difficulty predicting precise concentration values and tended to predict values closer to the middle of the target ranges.

The classification experiments for glucose and potassium further showed limited performance in distinguishing between multiple concentration ranges. Although potassium is known to affect ECG morphology\cite{teymouriECGFREQUE2022}, the experiments still showed difficulty reliably distinguishing the defined classes.

Overall, the additional experiments indicate that the applicability of ECG-based laboratory value prediction depends on the laboratory target and the relationship with the information available from the ECG and patient metadata. The predictive performance observed for NT-proBNP cannot be assumed to generalize directly to other laboratory values. Since the additional experiments partly used different preprocessing settings and feature configurations, the different laboratory values should not be compared directly.
\subsection{Limitations}
One limitation of this work is the use of data from a single clinical database. Consequently, the investigated patients and their associated data originated from Beth Israel Deaconess Medical Center\cite{PhysioNet-mimiciv-3.1, johnsonMIMICIVFR2023, pollardPHYSIONET2026}, indicating that the predictive performance may not generalize to patients from other clinical environments. External validation is required to confirm whether the predictive performance remains stable across different datasets.

Another limitation is the temporal matching between ECG recordings and laboratory measurements. In this work, ECG recordings were matched to the nearest NT-proBNP laboratory measurement using a temporal window of $\pm$3 hours. Since a time difference still remained for some ECG and laboratory measurement pairs, the ECG recording and the corresponding laboratory measurement may not necessarily reflect the same physiological state.

All primary experiments in this work used both ECG recordings and patient metadata, meaning that the actual influence of each modality has not been evaluated separately in this work. Although the comparison between 1-lead and 12-lead ECG recordings showed the influence of ECG lead configuration, the relative contributions of ECG and patient metadata remain unclear.
Furthermore, since the machine measurements used by both the 1-lead and 12-lead configurations were calculated on all 12 leads, the predictive performance difference may be underestimated.

Finally, another limitation is that the time difference between each ECG recording and its associated laboratory measurement was included as an input feature. This information may not be available when the model is used as a preliminary indicator of elevated NT-proBNP concentrations.
\subsection{Future Work}
Future research should first investigate the proposed approach using independent datasets to evaluate the generalization of the model. Applying the same prediction tasks and ECG configurations to unseen data would verify the stability of the results documented in this work. Independent datasets from different clinical environments, such as the Acutelines emergency department data-biobank\cite{wijkACUTELINES2026}, could provide opportunities for external validation.

Another direction of future research is a systematic comparison of each individual input modality for NT-proBNP concentration prediction and their contribution to the predictive performance. This research could also be extended to investigate whether the ECG lead configuration still improves the predictive performance when patient metadata is excluded from the model. Furthermore, alternative ECG representations and longer ECG recordings could be investigated. Recent work has shown that ECG-derived spectrograms combined with clinical information can support the prediction of patient deterioration in an emergency department setting\cite{vanwijkECGSPECTR2026}.

An additional extension would be a more systematic investigation of ECG lead selection. This work compared a 1-lead configuration using lead II with the complete 12-lead configuration, but did not investigate which leads or lead combinations contribute the most to the NT-proBNP prediction. This could identify whether similar or even better predictive performance can be achieved using fewer than 12 leads.

Another direction of future research could be the explainability of ECG-based laboratory value prediction. Explainability methods could be used to investigate which parts of the ECG waveform and which patient features contribute most to individual predictions\cite{lopezalcarazEXPLAINING2025}. This could also provide additional insights into the physiological relationships underlying ECG-based NT-proBNP prediction.

Lastly, future research could investigate other prediction formulations for three or more NT-proBNP concentration groups. The classes investigated in this work have a natural ordering from lower to higher NT-proBNP concentrations, while the current classification approach treats them as completely separate categorical classes. Therefore, future experiments could investigate an ordinal classification approach that considers the relationship between the classes.
\clearpage
\section{Conclusion}
This thesis investigated whether NT-proBNP concentrations can be predicted from ECG recordings and patient metadata using a multimodal deep learning framework. The experiments included regression, binary classification and three-class classification tasks and compared a 1-lead waveform configuration using lead II to a 12-lead waveform configuration. Furthermore, the physiological plausibility of ECG-based NT-proBNP prediction was investigated.

Across all investigated experiments, the 12-lead configuration achieved better predictive performance compared to the associated 1-lead configuration. In the regression experiment, the mean absolute error decreased from 534.98~pg/mL in the 1-lead configuration to 514.27~pg/mL in the 12-lead configuration. In the binary classification experiment, the AUROC increased from 0.849 to 0.876, while in the three-class classification experiment, the macro-averaged AUROC increased from 0.738 to 0.752. The three-class classification task showed lower predictive performance compared to the binary classification task. The 400--2000~pg/mL class showed the lowest proportion of correctly classified samples in both ECG configurations. Although the improvements were consistent, they remained moderate, indicating that the 1-lead configuration already provided considerable predictive information. The additional spatial information from the 12-lead configuration may have contributed to the improved prediction performance, while the regression error patterns remained similar.

The results in this work are consistent with the hypothesis that ECG signals contain indirect information about NT-proBNP concentration. Furthermore, elevated NT-proBNP concentrations may reflect underlying cardiovascular processes that are associated with altered ECG characteristics as discussed in Section~\ref{sec:physiological_interpretation}. The results should not be interpreted as showing that NT-proBNP directly influences ECG characteristics, but rather that NT-proBNP concentrations and ECG characteristics may both be associated with the same underlying physiological processes.

In conclusion, the findings in this work support the feasibility of ECG-based NT-proBNP prediction as a complementary source of information. Notably, the classification experiments showed that the model could distinguish between broader NT-proBNP concentration ranges, while regression tasks remain challenging. Therefore, the proposed approach may provide complementary information alongside direct laboratory measurements for NT-proBNP, while the additional experiments on hemoglobin, glucose and potassium indicate that the observed predictive performance cannot be directly generalized to other laboratory values. More research is required to determine whether the model can generalize to other patient populations and clinical settings.
\clearpage
\appendix
\section{Additional NT-proBNP experiments} \label{sec:additional_ntprobnp}
This appendix section presents additional experiments conducted for the prediction of NT-proBNP from ECG recordings and patient metadata. These experiments complement the primary NT-proBNP experiments presented in the main part of this thesis.
\subsection{Experimental Setup}
The additional experiments included an alternative classification formulation and comparisons between different regression configurations.

The general dataset construction and training pipeline followed the methodology described in the main part of this thesis unless specified otherwise. The model architecture remained unchanged across all additional experiments.

The first group of additional experiments investigated a classification task using simplified NT-proBNP concentration classes inspired by the structure of the New York Heart Association (NYHA) classification\cite{heidenreich2022AHAA2022} with a 1-lead ECG configuration. The NT-proBNP concentration ranges were below 300~pg/mL, between 300 and 1000~pg/mL, between 1000 and 2000~pg/mL, and above 2000~pg/mL. These ranges do not correspond directly to the NYHA classes. For these experiments, different loss functions were evaluated, including cross entropy, weighted cross entropy and weighted focal cross entropy loss. For the weighted configurations, class weights were determined based on the class distribution in the training set.

The next group of additional experiments investigated different target scaling settings using a 1-lead ECG configuration. The target scaling settings included MinMax and standard scaling, with mean squared error as the loss function. All configurations in this comparison used an earlier approach with a temporal matching window of one hour, an upper NT-proBNP clipping threshold of 10000~pg/mL and no patient vital signs.
\subsection{Results}
\subsubsection{NYHA Inspired Four-Class Classification}
\begin{table}[H]
\centering
\caption{Predictive results of the NYHA inspired NT-proBNP classification experiments.}
\label{tab:nyha_classification}
\small
\setlength{\tabcolsep}{4pt}
\begin{tabular}{lccccc}
\toprule
Loss configuration & Macro AUROC & 95\% CI & Accuracy & Sensitivity & F1 score \\
\midrule
Cross entropy
& 0.714 & 0.705--0.722 & 0.544 & 0.413 & 0.367 \\
Weighted cross entropy
& 0.726 & 0.717--0.734 & 0.471 & 0.413 & 0.398 \\
Weighted focal cross entropy
& 0.709 & 0.700--0.717 & 0.478 & 0.435 & 0.421 \\
\bottomrule
\end{tabular}
\end{table}

Table~\ref{tab:nyha_classification} shows the resulting macro-averaged AUROC values for the three evaluated loss configurations. The weighted cross entropy loss configuration achieved the highest macro-averaged AUROC of 0.726 (95\% CI: 0.717--0.734), followed by unweighted cross entropy with a macro-averaged AUROC of 0.714 (95\% CI: 0.705--0.722) and lastly weighted focal cross entropy loss with 0.709 (95\% CI: 0.700--0.717). The 95\% confidence intervals showed overlap across the three configurations. Cross entropy loss achieved the highest accuracy, while weighted focal cross entropy loss achieved the highest macro sensitivity and macro F1 score. 

\begin{figure}[H]
    \centering
    \begin{subfigure}[t]{0.65\textwidth}
        \centering
        \includegraphics[width=\textwidth]{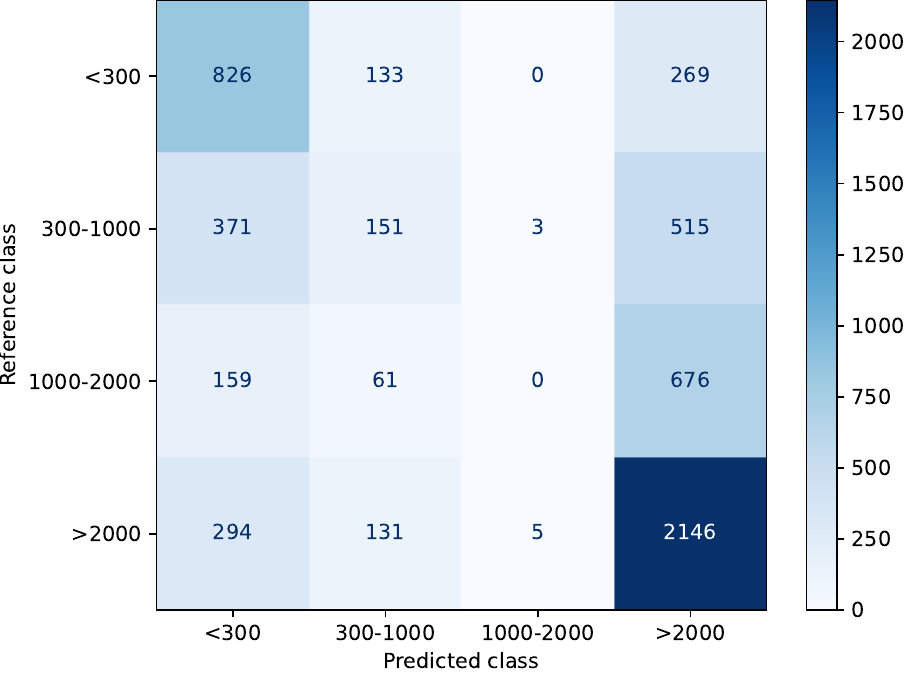}
        \caption{Unweighted cross entropy}
        \label{fig:nyha_ce}
    \end{subfigure}
    \begin{subfigure}[t]{0.65\textwidth}
        \centering
        \includegraphics[width=\textwidth]{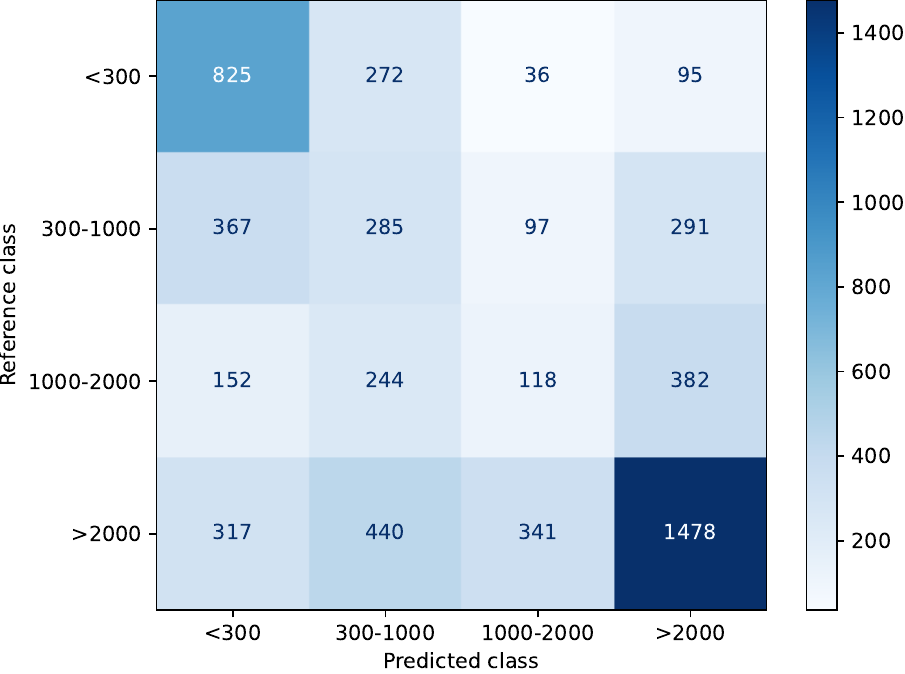}
        \caption{Weighted cross entropy}
        \label{fig:nyha_weighted_ce}
    \end{subfigure}
    \begin{subfigure}[t]{0.65\textwidth}
        \centering
        \includegraphics[width=\textwidth]{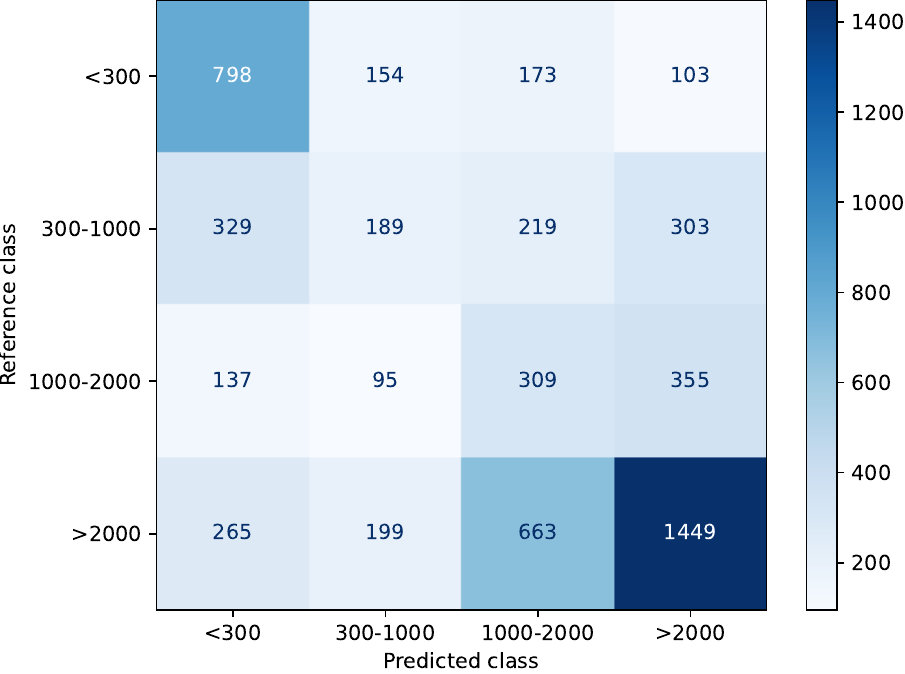}
        \caption{Weighted focal cross entropy}
        \label{fig:nyha_weighted_focal}
    \end{subfigure}
    \caption{Confusion matrices for the NYHA inspired NT-proBNP classifications}
    \label{fig:nyha_confusion_matrices}
\end{figure}

Figure~\ref{fig:nyha_confusion_matrices} shows the confusion matrices for the three evaluated configurations. The unweighted cross entropy configuration mostly assigned samples to the lowest and highest NT-proBNP classes, while the middle classes were rarely predicted. The weighted configurations showed a more distributed prediction pattern.

\subsubsection{Regression Configuration Experiments}
\begin{table}[H]
\centering
\caption{Predictive results of the additional NT-proBNP regression configurations.}
\label{tab:regression_configuration_experiments}
\begin{tabular}{lcc}
\toprule
Target scaling & MAE [pg/mL] & 95\% CI [pg/mL]\\
\midrule
MinMax   & 2390.46 & 2314.93--2460.34 \\
Standard & 2317.63 & 2242.62--2396.17 \\
\bottomrule
\end{tabular}
\end{table}
Table~\ref{tab:regression_configuration_experiments} summarizes the predictive performance of the evaluated regression configurations with standard scaling showing the smaller prediction error.

\begin{figure}[H]
    \centering

    \begin{subfigure}[t]{0.65\textwidth}
        \centering
        \includegraphics[width=\textwidth]{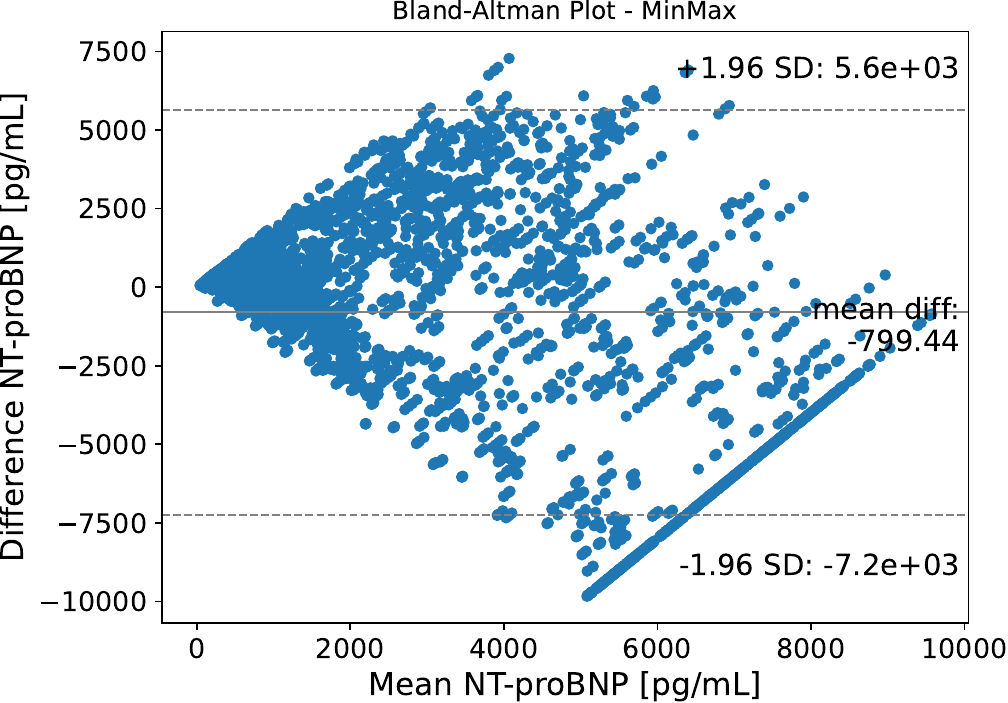}
        \caption{MinMax scaling}
        \label{fig:ba_minmax}
    \end{subfigure}

    \begin{subfigure}[t]{0.65\textwidth}
        \centering
        \includegraphics[width=\textwidth]{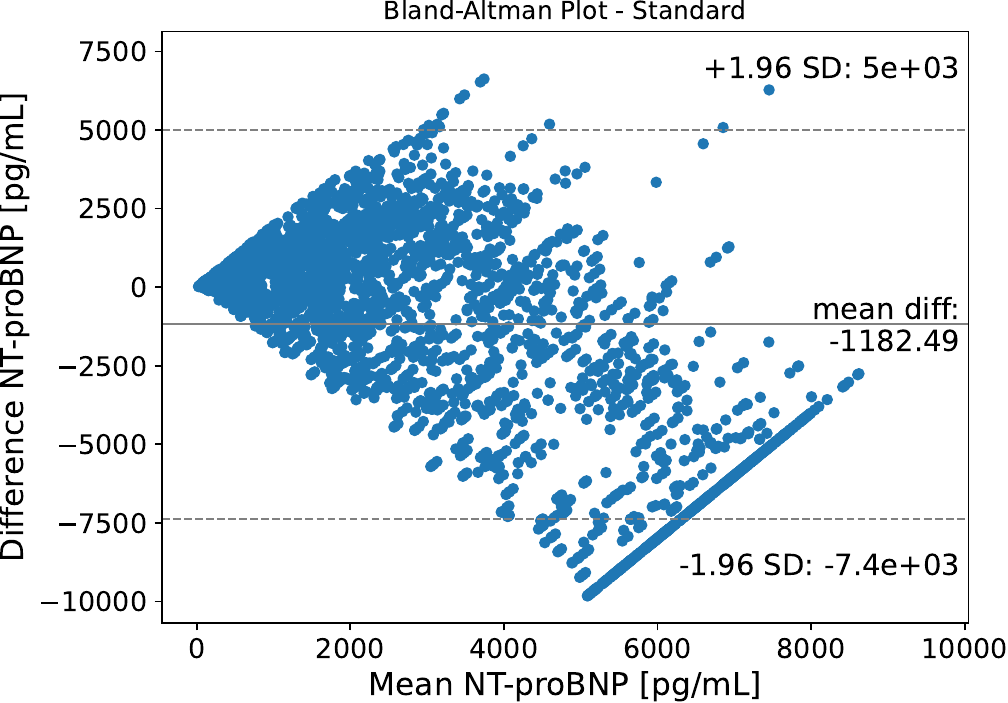}
        \caption{Standard scaling}
        \label{fig:ba_standard}
    \end{subfigure}

    \caption{Bland--Altman analysis of predicted and reference NT-proBNP concentrations for the MinMax and standard scaling configurations. The solid line represents the mean difference and the dashed lines indicate the 95\% limits of agreement.}
    \label{fig:ba_scaling_comparison}
\end{figure}

Figure~\ref{fig:ba_scaling_comparison} shows the Bland--Altman analysis for the target scaling configurations. MinMax scaling resulted in a mean difference of -799.44~pg/mL, while standard scaling showed a mean difference of -1182.49~pg/mL. The linear structure visible at high NT-proBNP concentrations resulted from the upper clipping threshold of 10000~pg/mL.

\begin{figure}[H]
    \centering
    \begin{subfigure}[t]{0.65\textwidth}
        \centering
        \includegraphics[width=\textwidth]{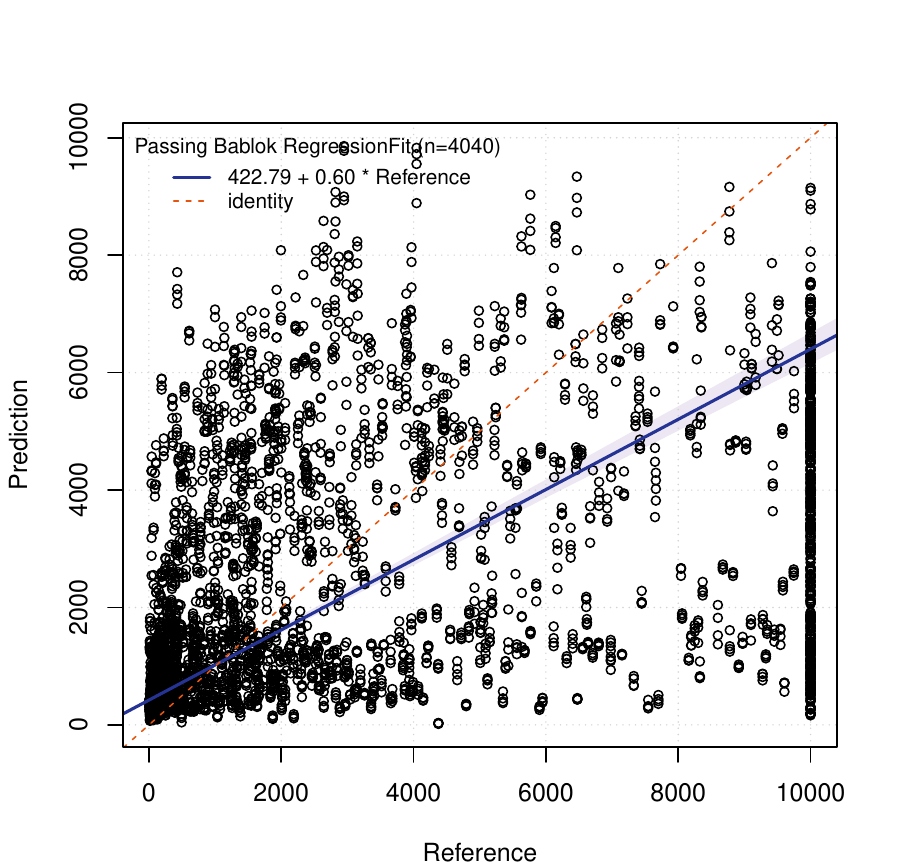}
        \caption{MinMax scaling}
        \label{fig:pb_minmax}
    \end{subfigure}
    \begin{subfigure}[t]{0.65\textwidth}
        \centering
        \includegraphics[width=\textwidth]{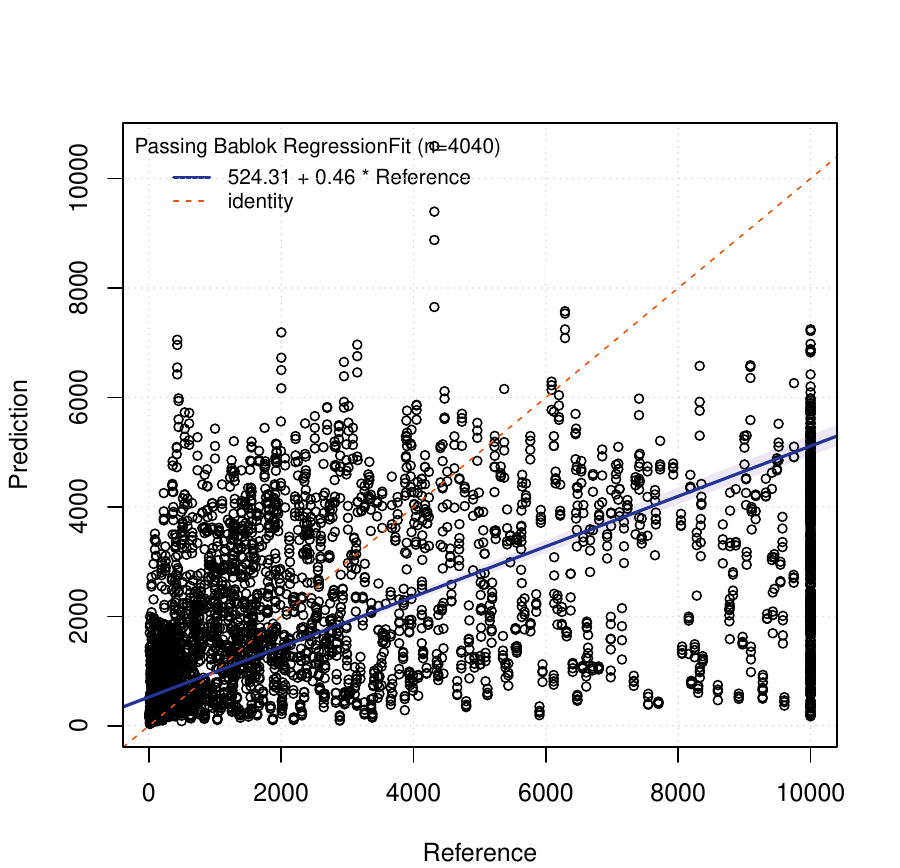}
        \caption{Standard scaling}
        \label{fig:pb_standard}
    \end{subfigure}
    \caption{Passing--Bablok regression analysis of predicted and reference NT-proBNP concentrations for the MinMax and standard scaling configurations. The solid line represents the Passing--Bablok regression fit, the dashed line represents the identity, and the shaded area represents the 95\% confidence interval.}
    \label{fig:pb_scaling_comparison}
\end{figure}

Figure~\ref{fig:pb_scaling_comparison} shows the Passing--Bablok regression analysis for both scaling configurations. MinMax scaling resulted in an intercept of 422.79~pg/mL (95\% CI: 400.79--454.43~pg/mL) and a slope of 0.6 (95\% CI: 0.57--0.63), while the standard scaling achieved an intercept of 524.31~pg/mL (95\% CI: 482.18--565.62~pg/mL) and a slope of 0.46 (95\% CI: 0.44--0.48). The linear structure at high NT-proBNP concentrations also resulted from the upper clipping threshold of 10000~pg/mL. 
\subsection{Interpretation}
The NYHA inspired classification experiments showed that different loss configurations affected the overall predictive performance of the model. Weighted cross entropy achieved the highest macro-averaged AUROC, while the overall confidence intervals showed overlap. Unweighted cross entropy predicted the lowest and highest NT-proBNP classes the most, whereas the 300--1000~pg/mL and 1000--2000~pg/mL classes were predicted the least. The weighted configurations showed better predictive results for the middle classes, while the overall predictive pattern appeared to be the same. These results also show the same difficulty for the model to predict multiple precise NT-proBNP concentration ranges, similar to the primary experiments.

The regression configuration experiments showed a small difference in mean absolute error between MinMax and standard target scaling. The Bland--Altman analysis showed a mean difference closer to 0 for the MinMax scaling compared to the standard target scaling, indicating a lower negative bias. The Passing--Bablok regression analysis further showed a slope closer to the identity for MinMax. Overall, the results indicate that standard scaling resulted in lower prediction error, while MinMax showed lower systematic bias. These findings are consistent with the regression pattern observed in the primary experiments, where precise NT-proBNP concentration prediction remained challenging.
\clearpage
\section{Hemoglobin} \label{sec:hemoglobin}
Hemoglobin was investigated as an additional laboratory target to assess whether the multimodal ECG-based prediction approach can be used for laboratory values other than NT-proBNP. The following experiments evaluate its predictive performance using ECG recordings and patient metadata. 
\subsection{Experimental Setup}
The experiments for hemoglobin followed the same methodology used for the primary NT-proBNP experiments. ECG recordings were combined with the same 33 features and processed using the same model architecture and training configuration.

Hemoglobin measurements were matched to ECG recordings using a temporal window of $\pm$2 minutes. The higher availability of hemoglobin measurements allowed for a smaller temporal window compared to the primary NT-proBNP measurements and resulted in 19176 ECG-laboratory pairs. No upper threshold was applied to the hemoglobin target values.

The target values were standardized using a \texttt{StandardScaler} fitted exclusively on the training split. L1 loss was used as the loss function and mean absolute error as the evaluation metric, identical to the primary NT-proBNP regression experiments in this work. Experiments were conducted for both 1-lead and 12-lead ECG configurations. 
\subsection{Results}
The test split consisted of 957 ECG recordings, corresponding to 3828 2.5 second ECG segments.

\begin{table}[H]
    \centering
    \caption{Mean absolute error with 95\% confidence intervals for hemoglobin prediction.}
    \label{tab:hemoglobin_regression}
    \begin{tabular}{lcc}
        \hline
        ECG configuration & MAE [g/dL] & 95\% CI [g/dL] \\
        \hline
        1-lead  & 1.6008 & 1.5585--1.6441 \\
        12-lead & 1.5991 & 1.5562--1.6413 \\
        \hline
    \end{tabular}
\end{table}

Table~\ref{tab:hemoglobin_regression} shows the regression performance of the 1-lead and 12-lead experiments. The 1-lead model achieved an MAE of 1.6008~g/dL (95\% CI: 1.5585--1.6441~g/dL), while the 12-lead model achieved an MAE of 1.5991~g/dL (95\% CI: 1.5562--1.6413~g/dL). 

\begin{figure}[H]
    \centering

    \begin{subfigure}[t]{0.65\textwidth}
        \centering
        \includegraphics[width=\textwidth]{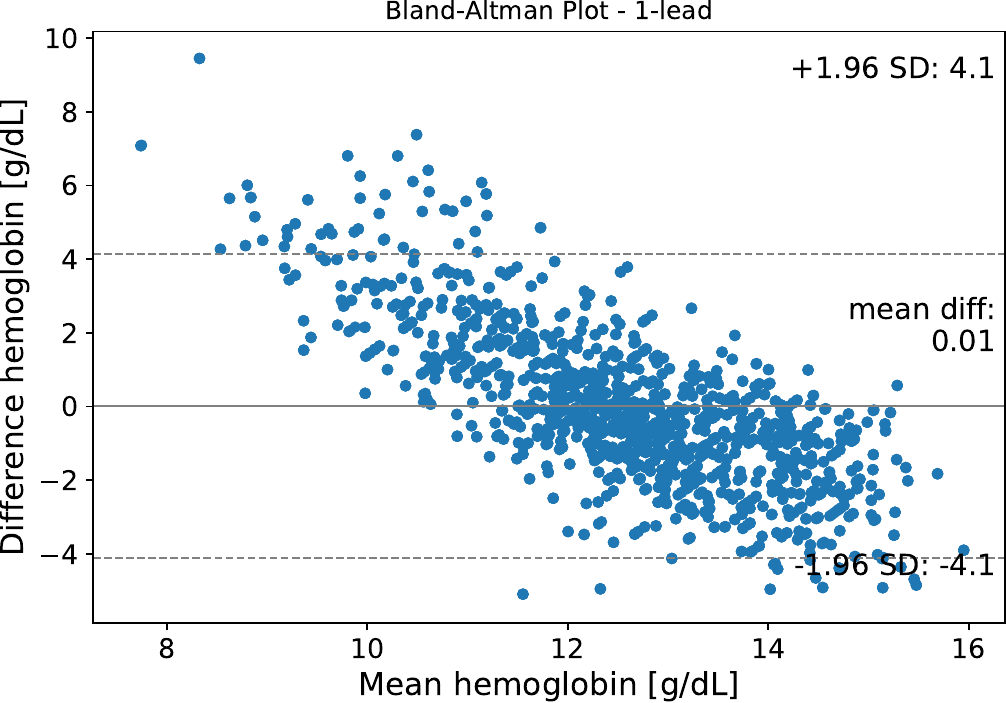}
        \caption{1-lead ECG}
        \label{fig:hemoglobin_ba_1lead}
    \end{subfigure}

    \begin{subfigure}[t]{0.65\textwidth}
        \centering
        \includegraphics[width=\textwidth]{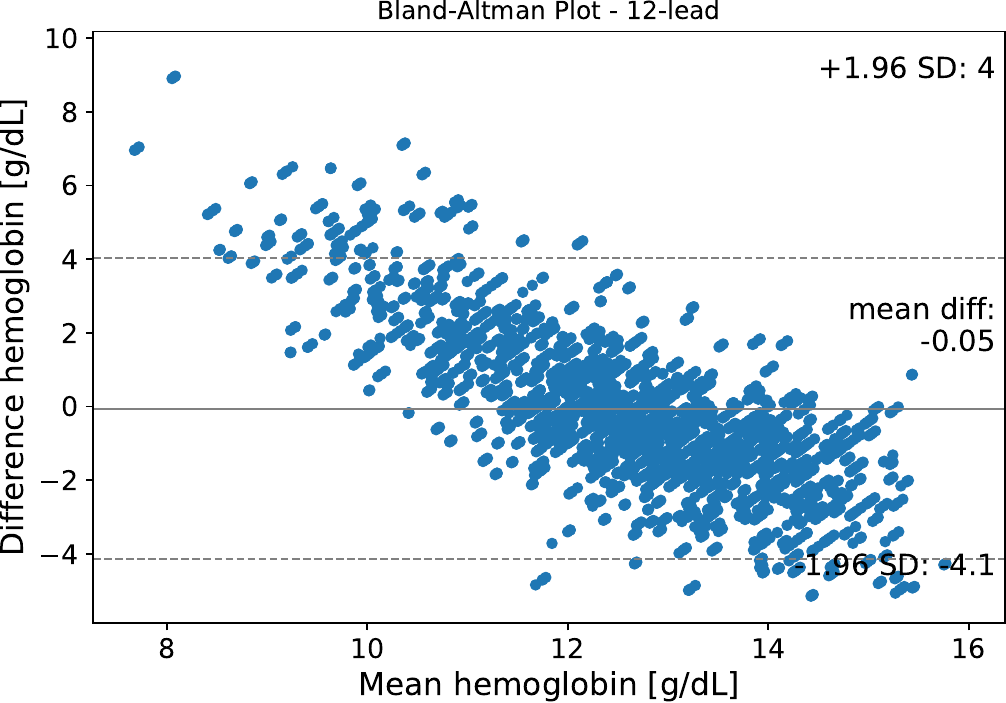}
        \caption{12-lead ECG}
        \label{fig:hemoglobin_ba_12lead}
    \end{subfigure}

    \caption{Bland--Altman analysis of predicted and reference hemoglobin concentrations for the 1-lead and 12-lead ECG configurations. The solid line represents the mean difference and the dashed lines indicate the 95\% limits of agreement.}
    \label{fig:hemoglobin_bland_altman}
\end{figure}

Figure~\ref{fig:hemoglobin_bland_altman} illustrates the Bland--Altman analysis for both ECG configurations. The 1-lead model showed a mean difference of 0.01~g/dL with limits of agreement from -4.12~g/dL to 4.14~g/dL, while the 12-lead model showed a mean difference of -0.05~g/dL with limits of agreement from -4.14~g/dL to 4.03~g/dL.

\begin{figure}[H]
    \centering
    \begin{subfigure}[t]{0.65\textwidth}
        \centering
        \includegraphics[width=\textwidth]{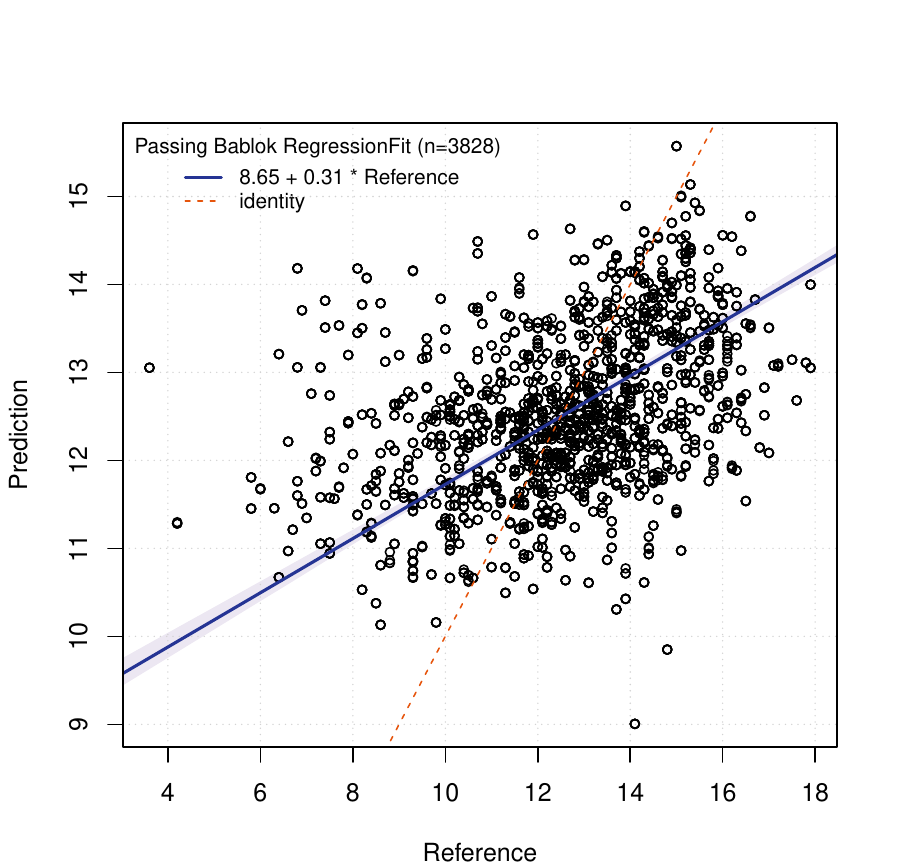}
        \caption{1-lead ECG}
        \label{fig:hemoglobin_pb_1lead}
    \end{subfigure}
    \begin{subfigure}[t]{0.65\textwidth}
        \centering
        \includegraphics[width=\textwidth]{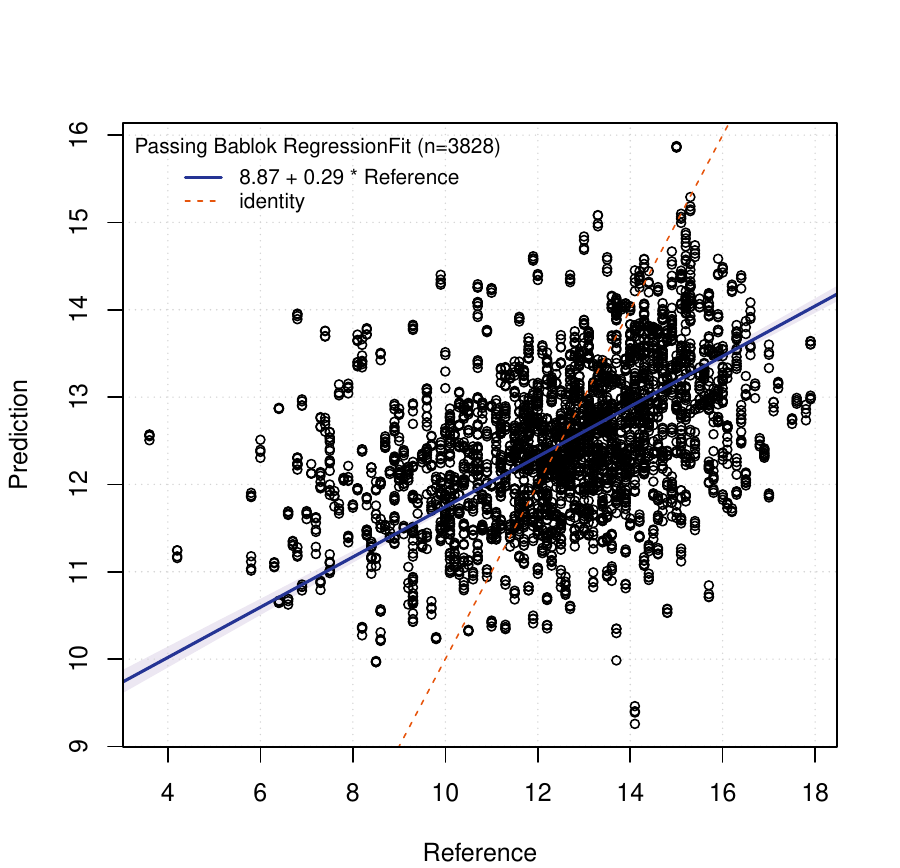}
        \caption{12-lead ECG}
        \label{fig:hemoglobin_pb_12lead}
    \end{subfigure}
    \caption{Passing--Bablok regression analysis of predicted and reference hemoglobin concentrations for the 1-lead and 12-lead ECG configurations. The solid line represents the Passing--Bablok regression fit, the dashed line represents the identity, and the shaded area represents the 95\% confidence interval.}
    \label{fig:hemoglobin_passing_bablok}
\end{figure}

Figure~\ref{fig:hemoglobin_passing_bablok} shows the Passing--Bablok regression analysis for 1-lead and 12-lead ECG configurations. The intercept for the 1-lead configuration was 8.65~g/dL (95\% CI: 8.47--8.87~g/dL) and the slope was 0.31 (95\% CI: 0.29--0.32), while for the 12-lead configuration the intercept was 8.87~g/dL (95\% CI: 8.70--9.05~g/dL) and the slope was 0.29 (95\% CI: 0.27--0.30).
\subsection{Interpretation}
The hemoglobin regression experiments showed very close predictive performance for the 1-lead and 12-lead ECG configurations with only minimal differences in mean absolute error between both ECG configurations.

The Bland--Altman analysis further showed that both configurations had very similar mean differences close to zero. However, the limits of agreement show that the predicted concentrations may differ considerably from the reference laboratory measurements.

Passing--Bablok analysis showed slopes clearly below 1 for both ECG configurations, indicating that the predicted hemoglobin concentrations varied considerably less than the corresponding reference values. Furthermore, the combination of intercept and slope indicates that the model tended to predict concentrations around the middle of the hemoglobin range.

Overall, the hemoglobin experiments showed no substantial difference in predictive performance using 1-lead compared to 12-lead ECG configurations. Furthermore, the regression task to predict precise hemoglobin concentrations remains challenging.
\clearpage
\section{Glucose} \label{sec:glucose}
Blood glucose was included as an additional prediction target to investigate the prediction of a metabolic laboratory value. The experiments document whether information associated with blood glucose concentration can be extracted from ECG recordings in combination with patient metadata.
\subsection{Experimental Setup}
The glucose experiments used an earlier approach with 21 features and a 1-lead ECG configuration similar to the experiments in Section~\ref{sec:potassium_exp_setup}. Patient vital signs were not included in the features. Values below 20~mg/dL and above 500~mg/dL were excluded in the preprocessing to limit the influence of extreme values.

The regression experiments used a temporal window of $\pm$5 minutes, where for each ECG recording the closest glucose measurement was selected. L1 loss was used as the loss function and mean absolute error as the evaluation metric. No target scaling was applied in this experiment.

Furthermore, additional regression configurations with a temporal window of $\pm$1 hour and different loss functions were investigated. These loss functions included Huber loss, Gaussian negative log-likelihood and a Clarke error grid weighted loss. These configurations were investigated to determine whether different loss functions could improve the prediction of blood glucose concentrations.

Finally, a three-class classification task for glucose was investigated. This experiment used the same $\pm$5 minute temporal window. The classes were defined as hypoglycemia for glucose concentrations below 70~mg/dL, hyperglycemia for concentrations above 180~mg/dL and glucose concentrations between these thresholds\cite{battelinoCLINICALT2019}. Weighted focal cross entropy loss with class weights accounting for the unequal class distribution was used, similar to the other classification tasks in this work.
\subsection{Results}
The test split consisted of 1797 ECG recordings, corresponding to 7188 2.5 second ECG segments for the datasets with temporal windows of $\pm$5 minutes.
\subsubsection{Regression}
The regression experiment using a temporal window of $\pm$5 minutes achieved a mean absolute error of 34.52~mg/dL with a 95\% confidence interval from 33.47~mg/dL to 35.60~mg/dL.

\begin{table}[H]
\centering
\caption{Regression configurations for glucose prediction.}
\label{tab:glucose_regression_configs}
\begin{tabular}{lcc}
\hline
Loss configuration & MAE [mg/dL] & 95\% CI [mg/dL]\\
\hline
L1 loss & 34.26 & 33.81--34.69 \\
Huber loss & 34.32 & 33.87--34.81 \\
Gaussian negative log-likelihood & 37.75 & 37.35--38.14 \\
Weighted Clarke error grid loss & 48.40 & 48.04--48.72 \\
\hline
\end{tabular}
\end{table}

Table~\ref{tab:glucose_regression_configs} shows the predictive performance of the regression experiments for blood glucose using a $\pm$1 hour temporal window and different loss functions. L1 loss achieved the lowest mean absolute error of 34.26~mg/dL (95\% CI: 33.81--34.69~mg/dL), while weighted Clarke error grid loss achieved the highest error of 48.40~mg/dL (95\% CI: 48.04--48.72~mg/dL).

\begin{figure}[H]
    \centering
    \includegraphics[width=0.65\textwidth]{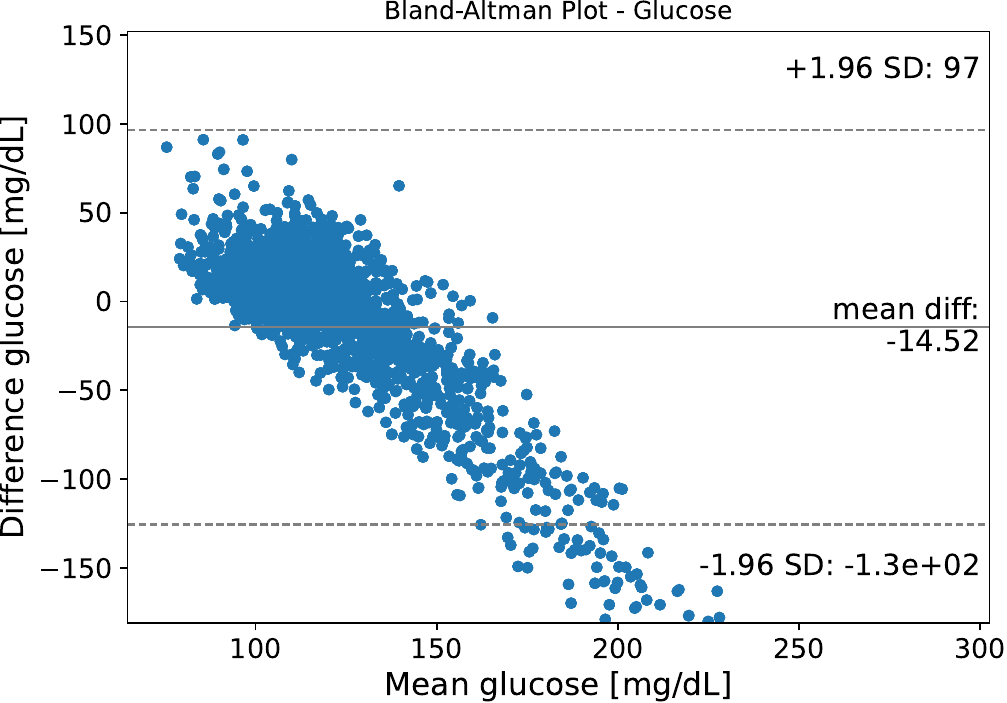}
    \caption{Bland--Altman analysis of predicted and reference glucose concentrations using the $\pm$5 minute temporal window. The solid line represents the mean difference and the dashed lines indicate the 95\% limits of agreement.}
    \label{fig:glucose_bland_altman}
\end{figure}

Figure~\ref{fig:glucose_bland_altman} shows the Bland--Altman analysis for the glucose prediction using the $\pm$5 minute temporal window. The mean difference was -14.52~mg/dL, with limits of agreement from -125.56~mg/dL to 96.51~mg/dL.

\begin{figure}[H]
    \centering
    \includegraphics[width=0.65\textwidth]{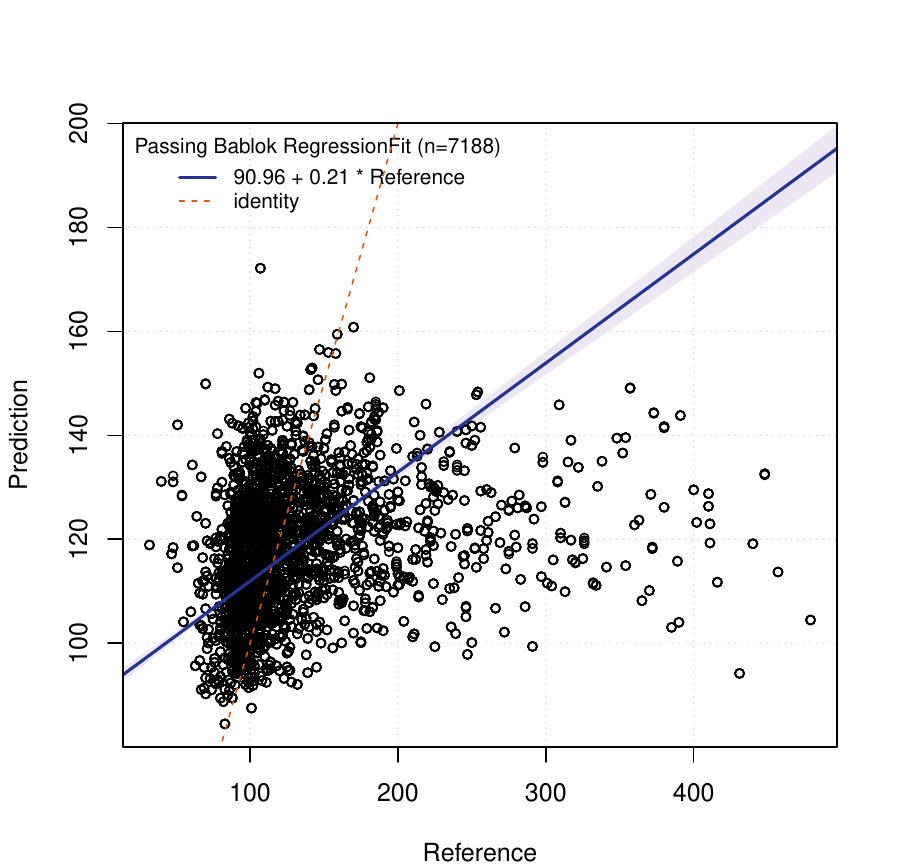}
    \caption{Passing--Bablok regression analysis for glucose prediction.}
    \label{fig:glucose_passing_bablok}
\end{figure}

Figure~\ref{fig:glucose_passing_bablok} shows the Passing--Bablok regression analysis for glucose. The intercept was 90.96~mg/dL (95\% CI: 89.56--92.47~mg/dL), while the slope was 0.21 (95\% CI: 0.20--0.22).

\subsubsection{Classification}
\begin{table}[H]
\centering
\caption{Three-class classification performance for glucose prediction.}
\label{tab:glucose_classification}
\begin{tabular}{lccccc}
\hline
ECG configuration & Macro AUROC & 95\% CI & Accuracy & Sensitivity & F1 score \\
\hline
1-lead & 0.632 & 0.610--0.652 & 0.554 & 0.420 & 0.338 \\
\hline
\end{tabular}
\end{table}

Table~\ref{tab:glucose_classification} shows the classification performance for glucose prediction. The model achieved a macro-averaged AUROC of 0.632 (95\% CI: 0.610--0.652) and an accuracy of 0.554. The sensitivity and F1 score achieved were less than 0.5.

\begin{figure}[H]
    \centering
    \includegraphics[width=0.65\textwidth]{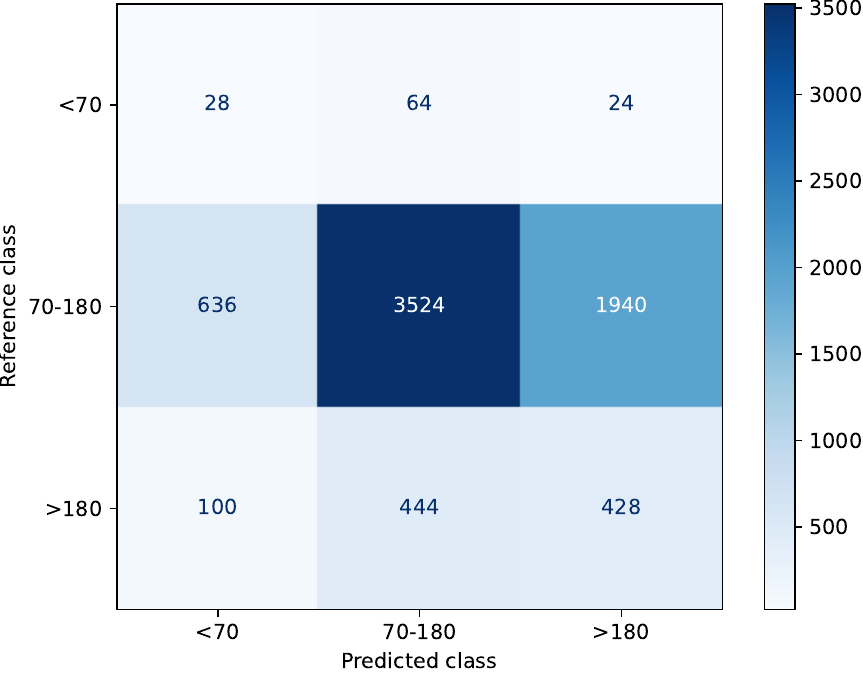}
    \caption{Confusion matrix for the three-class glucose classification experiment.}
    \label{fig:glucose_confusion_matrix}
\end{figure}

Figure~\ref{fig:glucose_confusion_matrix} shows the confusion matrix for the glucose classification experiment. The 70--180~mg/dL class was predicted most frequently, while the other classes were predicted considerably less frequently.
\subsection{Interpretation}
The Bland--Altman analysis showed wide limits of agreement, indicating that the predicted glucose concentration may differ considerably from the reference glucose concentration.

Passing--Bablok further showed a slope considerably below the identity slope of 1. Combined with the intercept, this supports the pattern that the predicted glucose concentrations varied less than the reference concentrations. In addition, the regression experiment indicated that the model tended to predict values around the middle of the glucose range, similar to the hemoglobin experiments.

The comparison between the different loss functions for the glucose regression experiment showed that L1 loss achieved similar predictive performance to Huber loss. The other loss functions resulted in higher mean absolute error and therefore did not improve the predictive performance.

The three-class classification showed similar results to the regression experiment. The middle class was predicted most frequently, while the other classes were predicted considerably less reliably.

Overall, the glucose experiments showed limited predictive performance, with difficulty in predicting precise glucose concentrations and classifying broader concentration ranges.
\clearpage
\section{Potassium} \label{sec:potassium}
Potassium was selected as an additional laboratory target since changes in potassium concentration are known to affect cardiac electrophysiology and ECG morphology\cite{teymouriECGFREQUE2022}. The following experiments document the use of ECG recordings and patient metadata for the prediction of potassium.
\subsection{Experimental Setup} \label{sec:potassium_exp_setup}
The potassium experiments used an earlier approach with a 1-lead ECG configuration combined with 21 features. A temporal window of $\pm$5 minutes was used for the matching of ECG recordings to laboratory measurements. Potassium values outside the range from 1.5~mEq/L to 8.0~mEq/L were excluded during preprocessing to reduce the influence of extreme outliers. Furthermore, patient vital signs were not included in the metadata.

The experiments included both a regression experiment and a three-class classification experiment. For the regression experiment, L1 loss was used as the loss function and mean absolute error as the evaluation metric. No target scaling was applied for this experiment.

The three classes were defined as low, normal and elevated potassium concentrations. The thresholds were concentrations below 3.5~mEq/L, between 3.5~mEq/L and 5.0~mEq/L and above 5.0~mEq/L\cite{kimPOTASSIUM}. Weighted focal cross entropy loss with class weights to account for the class distribution was used.
\subsection{Results}
The test split for these experiments consisted of 1872 ECG recordings, corresponding to 7488 2.5 second ECG segments.
\subsubsection{Regression}
The model achieved a mean absolute error of 0.51~mEq/L with a 95\% confidence interval from 0.50~mEq/L to 0.52~mEq/L.

\begin{figure}[H]
    \centering
    \includegraphics[width=0.65\textwidth]{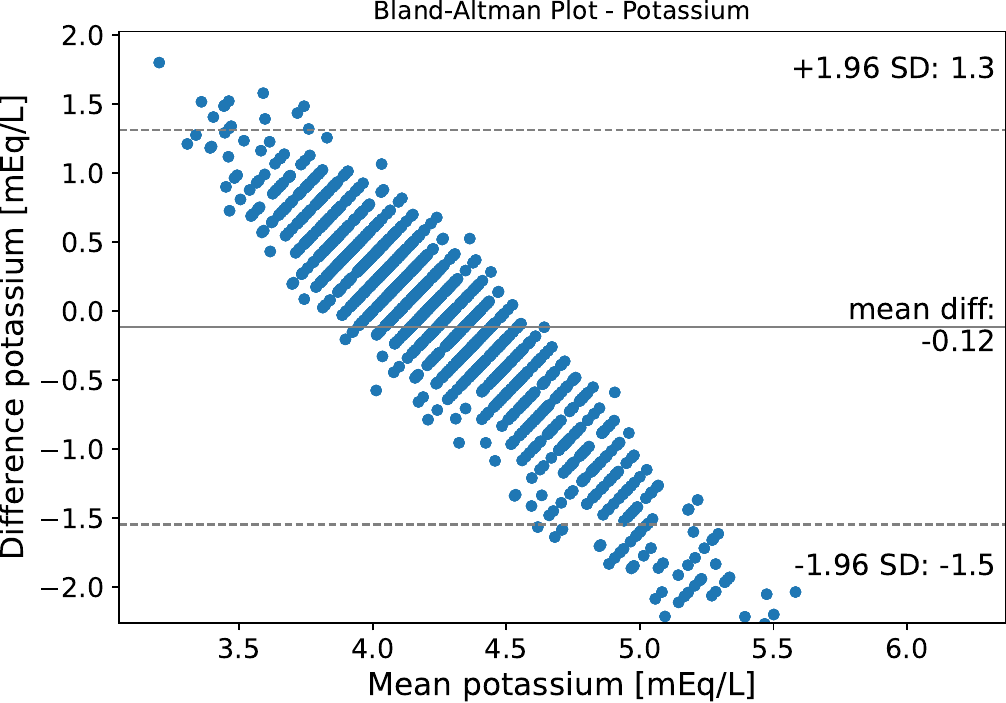}
    \caption{Bland--Altman analysis of predicted and reference potassium concentrations using the 1-lead ECG configuration. The solid line represents the mean difference and the dashed lines indicate the 95\% limits of agreement.}
    \label{fig:potassium_bland_altman}
\end{figure}

Figure~\ref{fig:potassium_bland_altman} shows the Bland--Altman analysis for the potassium prediction. The mean difference was -0.12~mEq/L with limits of agreement from -1.5~mEq/L to 1.3~mEq/L.

\begin{figure}[H]
    \centering
    \includegraphics[width=0.65\textwidth]{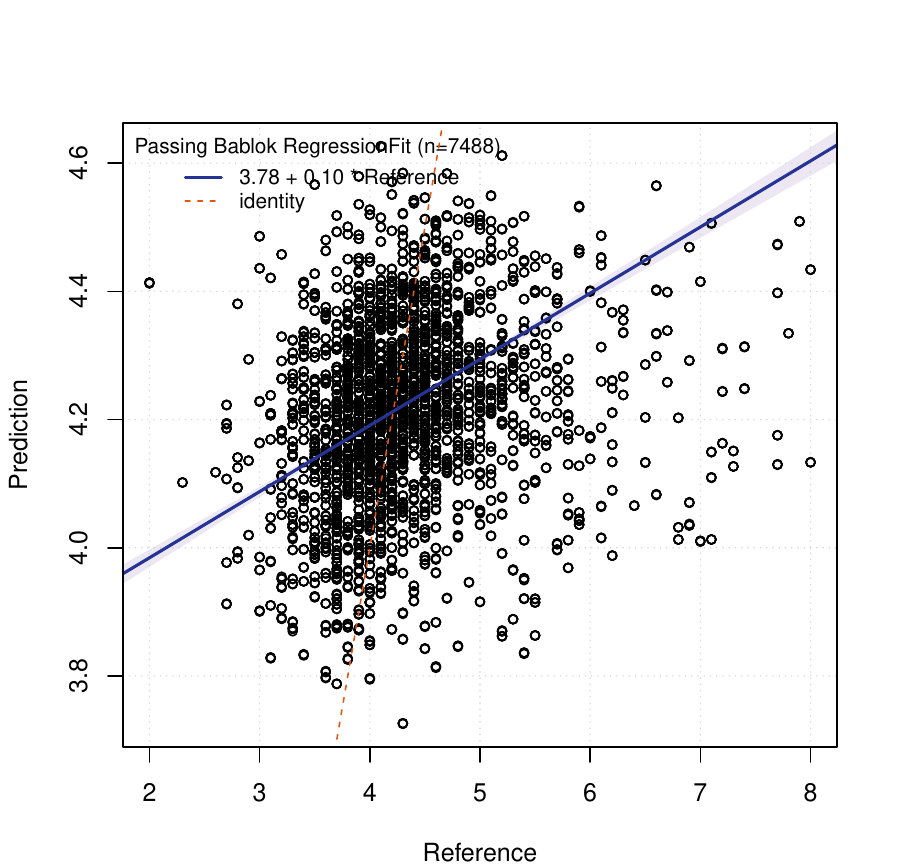}
    \caption{Passing--Bablok regression analysis for potassium prediction.}
    \label{fig:potassium_passing_bablok}
\end{figure}

Figure~\ref{fig:potassium_passing_bablok} shows the Passing--Bablok regression analysis for potassium concentration prediction with an intercept of 3.78~mEq/L (95\% CI: 3.75--3.80~mEq/L) and a slope of 0.10 (95\% CI: 0.10--0.11).
\subsubsection{Classification}
\begin{table}[H]
\centering
\caption{Three-class classification performance for potassium prediction.}
\label{tab:potassium_classification}
\begin{tabular}{lccccc}
\hline
ECG configuration & Macro AUROC & 95\% CI & Accuracy & Sensitivity & F1 score \\
\hline
1-lead & 0.629 & 0.617--0.642 & 0.488 & 0.444 & 0.356 \\
\hline
\end{tabular}
\end{table}

Table~\ref{tab:potassium_classification} shows the predictive performance of the three-class classification experiment for potassium. A macro-averaged AUROC of 0.629 (95\% CI: 0.617--0.642) was achieved, while accuracy, sensitivity and F1 score were all below 0.5.

\begin{figure}[H]
    \centering
    \includegraphics[width=0.65\textwidth]{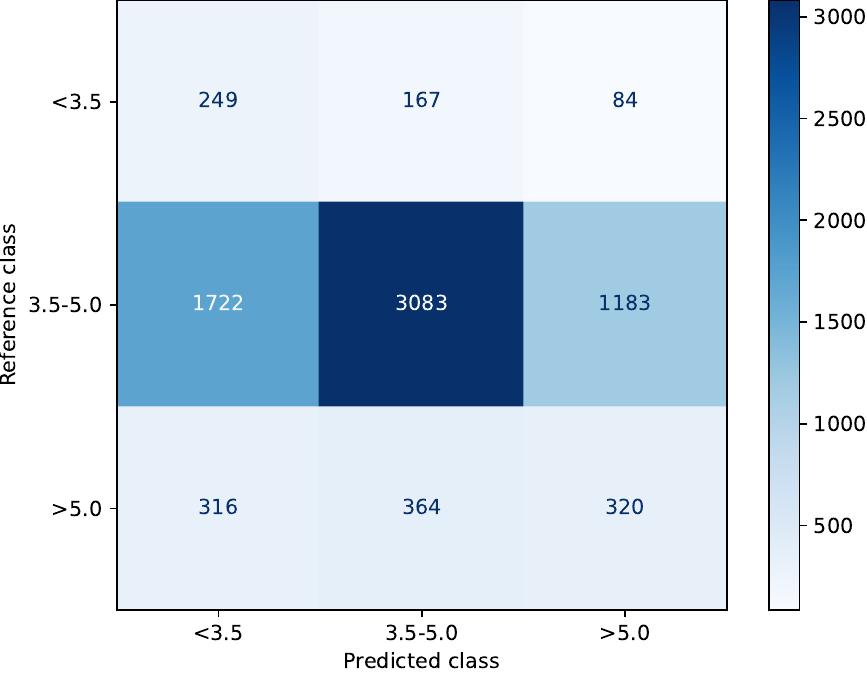}
    \caption{Confusion matrix for the three-class potassium classification experiment.}
    \label{fig:potassium_confusion_matrix}
\end{figure}

Figure~\ref{fig:potassium_confusion_matrix} shows that the normal potassium class was predicted most frequently. The low and elevated potassium classes were frequently misclassified.
\subsection{Interpretation}
For the regression experiment, the Bland--Altman analysis showed wide limits of agreement, indicating that the predicted concentrations may differ considerably from the corresponding laboratory measurements.

Passing--Bablok analysis further supported this pattern. The slope was 0.1, which is considerably below the identity slope of 1. Combined with the intercept, this indicates that the model had considerable difficulty predicting precise potassium concentrations.

The three-class classification experiment showed the same pattern in Figure~\ref{fig:potassium_confusion_matrix}, where the predicted classes varied considerably compared to the reference classes.

The potassium experiments showed limited predictive performance, with difficulty predicting precise potassium concentrations and classifying broader concentration ranges, similar to glucose.
\clearpage
\printbibliography

@article{alcarazhancingclinicaldecision2025,
  title = {Enhancing Clinical Decision Support with Physiological Waveforms — {{A}} Multimodal Benchmark in Emergency Care},
  author = {Alcaraz, Juan Miguel Lopez and Bouma, Hjalmar and Strodthoff, Nils},
  date = {2025-06-01},
  journaltitle = {Computers in Biology and Medicine},
  shortjournal = {Computers in Biology and Medicine},
  volume = {192},
  pages = {110196},
  issn = {0010-4825},
  doi = {10.1016/j.compbiomed.2025.110196},
  url = {https://www.sciencedirect.com/science/article/pii/S0010482525005475},
  urldate = {2026-04-16}
}

@article{alcaraznormalitypredictionforecasting2025,
  title = {Abnormality Prediction and Forecasting of Laboratory Values from Electrocardiogram Signals Using Multimodal Deep Learning},
  author = {Alcaraz, Juan Miguel Lopez and Strodthoff, Nils},
  date = {2025-11-10},
  journaltitle = {Scientific Reports},
  shortjournal = {Sci Rep},
  volume = {15},
  number = {1},
  pages = {39362},
  publisher = {Nature Publishing Group},
  issn = {2045-2322},
  doi = {10.1038/s41598-025-26715-5},
  url = {https://www.nature.com/articles/s41598-025-26715-5},
  urldate = {2026-04-16},
  langid = {english}
}

@article{habehhMACHINELE2021,
  title = {Machine {{Learning}} in {{Healthcare}}},
  author = {Habehh, Hafsa and Gohel, Suril},
  date = {2021-12-16},
  journaltitle = {Current Genomics},
  shortjournal = {Curr Genomics},
  volume = {22},
  number = {4},
  eprint = {35273459},
  eprinttype = {pubmed},
  pages = {291--300},
  issn = {1389-2029},
  doi = {10.2174/1389202922666210705124359},
  url = {https://pmc.ncbi.nlm.nih.gov/articles/PMC8822225/},
  urldate = {2026-04-16},
  pmcid = {PMC8822225}
}

@article{klineMULTIMODAL2022,
  title = {Multimodal Machine Learning in Precision Health: {{A}} Scoping Review},
  shorttitle = {Multimodal Machine Learning in Precision Health},
  author = {Kline, Adrienne and Wang, Hanyin and Li, Yikuan and Dennis, Saya and Hutch, Meghan and Xu, Zhenxing and Wang, Fei and Cheng, Feixiong and Luo, Yuan},
  date = {2022-11-07},
  journaltitle = {npj Digital Medicine},
  shortjournal = {npj Digit. Med.},
  volume = {5},
  number = {1},
  pages = {171},
  publisher = {Nature Publishing Group},
  issn = {2398-6352},
  doi = {10.1038/s41746-022-00712-8},
  url = {https://www.nature.com/articles/s41746-022-00712-8},
  urldate = {2026-04-16},
  langid = {english}
}

@article{marzoogMACHINELE2026,
  title = {Machine Learning Detection of Metabolic Syndrome through Single-Lead Electrocardiography: {{Digital}} Phenotyping of Cardiac Electrical Alterations},
  shorttitle = {Machine Learning Detection of Metabolic Syndrome through Single-Lead Electrocardiography},
  author = {Marzoog, Basheer Abdulah and Kopylov, Philipp},
  date = {2026-05-01},
  journaltitle = {Biomedical Signal Processing and Control},
  shortjournal = {Biomedical Signal Processing and Control},
  volume = {116},
  pages = {109616},
  issn = {1746-8094},
  doi = {10.1016/j.bspc.2026.109616},
  url = {https://www.sciencedirect.com/science/article/pii/S1746809426001709},
  urldate = {2026-04-16}
}

@article{patelMACHINELE2015,
  title = {Machine Learning Approaches for Integrating Clinical and Imaging Features in Late-Life Depression Classification and Response Prediction},
  author = {Patel, Meenal J. and Andreescu, Carmen and Price, Julie C. and Edelman, Kathryn L. and Reynolds III, Charles F. and Aizenstein, Howard J.},
  date = {2015},
  journaltitle = {International Journal of Geriatric Psychiatry},
  volume = {30},
  number = {10},
  pages = {1056--1067},
  issn = {1099-1166},
  doi = {10.1002/gps.4262},
  url = {https://onlinelibrary.wiley.com/doi/abs/10.1002/gps.4262},
  urldate = {2026-04-16},
  langid = {english}
}

@article{ceasovschihELECTROCAR2024,
  title = {Electrocardiogram {{Features}} in {{Non-Cardiac Diseases}}: {{From Mechanisms}} to {{Practical Aspects}}},
  shorttitle = {Electrocardiogram {{Features}} in {{Non-Cardiac Diseases}}},
  author = {Ceasovschih, Alexandr and Șorodoc, Victorița and Covantsev, Serghei and Balta, Anastasia and Uzokov, Jamol and Kaiser, Sergio E and Almaghraby, Abdallah and Lionte, Cătălina and Stătescu, Cristian and Sascău, Radu A and Onofrei, Viviana and Haliga, Raluca Ecaterina and Stoica, Alexandra and Bologa, Cristina and Ailoaei, Ștefan and Şener, Yusuf Ziya and Kounis, Nicholas G and Șorodoc, Laurențiu},
  date = {2024-04-20},
  journaltitle = {Journal of Multidisciplinary Healthcare},
  volume = {17},
  eprint = {38659633},
  eprinttype = {pubmed},
  pages = {1695--1719},
  publisher = {Dove Medical Press},
  issn = {null},
  doi = {10.2147/JMDH.S445549},
  url = {https://www.tandfonline.com/doi/abs/10.2147/JMDH.S445549},
  urldate = {2026-05-05}
}

@article{lecunDEEPLEARN2015,
  title = {Deep Learning},
  author = {LeCun, Yann and Bengio, Yoshua and Hinton, Geoffrey},
  date = {2015-05},
  journaltitle = {Nature},
  volume = {521},
  number = {7553},
  pages = {436--444},
  publisher = {Nature Publishing Group},
  issn = {1476-4687},
  doi = {10.1038/nature14539},
  url = {https://www.nature.com/articles/nature14539},
  urldate = {2026-05-12},
  langid = {english}
}

@article{gallowayDEVELOPMEN2019,
  title = {Development and {{Validation}} of a {{Deep-Learning Model}} to {{Screen}} for {{Hyperkalemia From}} the {{Electrocardiogram}}},
  author = {Galloway, Conner D. and Valys, Alexander V. and Shreibati, Jacqueline B. and Treiman, Daniel L. and Petterson, Frank L. and Gundotra, Vivek P. and Albert, David E. and Attia, Zachi I. and Carter, Rickey E. and Asirvatham, Samuel J. and Ackerman, Michael J. and Noseworthy, Peter A. and Dillon, John J. and Friedman, Paul A.},
  date = {2019-05-01},
  journaltitle = {JAMA Cardiology},
  shortjournal = {JAMA Cardiol.},
  volume = {4},
  number = {5},
  pages = {428--436},
  issn = {2380-6583},
  doi = {10.1001/jamacardio.2019.0640},
  url = {https://doi.org/10.1001/jamacardio.2019.0640},
  urldate = {2026-07-17}
}

@article{prajapatiCOMPREHENS2026,
  title = {A Comprehensive Review on {{ECG}} Diagnosis and Report Generation Using {{AI}}: {{Technologies}}, Challenges, and Future Research Directions},
  shorttitle = {A Comprehensive Review on {{ECG}} Diagnosis and Report Generation Using {{AI}}},
  author = {Prajapati, Varshil and Kolhar, Shrikrishna and Barbole, Dhanashree},
  date = {2026-07-01},
  journaltitle = {Informatics in Medicine Unlocked},
  shortjournal = {Inform. Med. Unlocked},
  volume = {64},
  pages = {101777},
  issn = {2352-9148},
  doi = {10.1016/j.imu.2026.101777},
  url = {https://www.sciencedirect.com/science/article/pii/S235291482600047X},
  urldate = {2026-07-21}
}

@article{sauEMERGIN2023,
  title = {\textbf{­­} {{The}} Emerging Role of Artificial Intelligence Enabled Electrocardiograms in Healthcare},
  author = {Sau, Arunashis and Ng, Fu Siong},
  date = {2023-07},
  journaltitle = {BMJ Medicine},
  shortjournal = {BMJ Med.},
  volume = {2},
  number = {1},
  pages = {e000193},
  issn = {2754-0413},
  doi = {10.1136/bmjmed-2022-000193},
  url = {https://bmjmedicine.bmj.com/lookup/doi/10.1136/bmjmed-2022-000193},
  urldate = {2026-07-21},
  langid = {english}
}

@article{kraikMOSTCOMMO2025,
  title = {The Most Common Errors in Automatic {{ECG}} Interpretation},
  author = {Kraik, Krzysztof and Dykiert, Irena Anna and Niewiadomska, Joanna and Ziemer-Szymańska, Marta and Mikołajczak, Karolina and Kreń, Mikołaj and Kukiełka, Piotr and Martuszewski, Adrian and Harych, Tomasz and Poręba, Rafał and Gać, Paweł and Poręba, Małgorzata},
  date = {2025-05-22},
  journaltitle = {Frontiers in Physiology},
  shortjournal = {Front. Physiol.},
  volume = {16},
  publisher = {Frontiers},
  issn = {1664-042X},
  doi = {10.3389/fphys.2025.1590170},
  url = {https://www.frontiersin.org/journals/physiology/articles/10.3389/fphys.2025.1590170/full},
  urldate = {2026-07-21},
  langid = {english}
}

@article{martinez-sellesCURRENTFU2023,
  title = {Current and {{Future Use}} of {{Artificial Intelligence}} in {{Electrocardiography}}},
  author = {Martínez-Sellés, Manuel and Marina-Breysse, Manuel},
  date = {2023-04},
  journaltitle = {Journal of Cardiovascular Development and Disease},
  shortjournal = {J. Cardiovasc. Dev. Dis.},
  volume = {10},
  number = {4},
  pages = {175},
  publisher = {Multidisciplinary Digital Publishing Institute},
  issn = {2308-3425},
  doi = {10.3390/jcdd10040175},
  url = {https://www.mdpi.com/2308-3425/10/4/175},
  urldate = {2026-07-21},
  langid = {english}
}

@book{iaizzoHANDBOOKC2024,
  title = {Handbook of {{Cardiac Anatomy}}, {{Physiology}}, and {{Devices}}},
  author = {Iaizzo, Paul A.},
  date = {2024},
  publisher = {Springer},
  location = {Cham, SWITZERLAND},
  url = {http://ebookcentral.proquest.com/lib/zbmed-ebooks/detail.action?docID=31823687},
  urldate = {2026-07-22},
  isbn = {978-3-031-72581-4}
}

@article{yandleBTYPENATR2015,
  title = {B-Type {{Natriuretic Peptide}} Circulating Forms: {{Analytical}} and Bioactivity Issues},
  shorttitle = {B-Type {{Natriuretic Peptide}} Circulating Forms},
  author = {Yandle, Tim G. and Richards, A. Mark},
  date = {2015-08-25},
  journaltitle = {Clinica Chimica Acta},
  shortjournal = {Clin. Chim. Acta},
  volume = {448},
  pages = {195--205},
  issn = {0009-8981},
  doi = {10.1016/j.cca.2015.07.004},
  url = {https://www.sciencedirect.com/science/article/pii/S0009898115003290},
  urldate = {2026-07-23}
}

@article{nishikimiBTYPENATR2022,
  title = {B-{{Type Natriuretic Peptide}} ({{BNP}}) {{Revisited}}—{{Is BNP Still}} a {{Biomarker}} for {{Heart Failure}} in the {{Angiotensin Receptor}}/{{Neprilysin Inhibitor Era}}?},
  author = {Nishikimi, Toshio and Nakagawa, Yasuaki},
  date = {2022-07},
  journaltitle = {Biology},
  volume = {11},
  number = {7},
  pages = {1034},
  publisher = {Multidisciplinary Digital Publishing Institute},
  issn = {2079-7737},
  doi = {10.3390/biology11071034},
  url = {https://www.mdpi.com/2079-7737/11/7/1034},
  urldate = {2026-07-23},
  langid = {english}
}

@article{oremusSYSTEMATIC2014,
  title = {A Systematic Review of {{BNP}} and {{NT-proBNP}} in the Management of Heart Failure: Overview and Methods},
  shorttitle = {A Systematic Review of {{BNP}} and {{NT-proBNP}} in the Management of Heart Failure},
  author = {Oremus, Mark and McKelvie, Robert and Don-Wauchope, Andrew and Santaguida, Pasqualina L. and Ali, Usman and Balion, Cynthia and Hill, Stephen and Booth, Ronald and Brown, Judy A. and Bustamam, Amy and Sohel, Nazmul and Raina, Parminder},
  date = {2014-08-01},
  journaltitle = {Heart Failure Reviews},
  shortjournal = {Heart Fail. Rev.},
  volume = {19},
  number = {4},
  pages = {413--419},
  issn = {1573-7322},
  doi = {10.1007/s10741-014-9440-0},
  url = {https://doi.org/10.1007/s10741-014-9440-0},
  urldate = {2026-07-30},
  langid = {english}
}

@article{minamisawa2023UPDAT2025,
  title = {2023 {{Update}} of the {{Japanese Heart Failure Society Scientific Statement}} on {{BNP}} and {{NT-proBNP Levels}} in {{Heart Failure Practice}}},
  author = {family=Minamisawa, given=MASATOSHI, given-i=MASATOSHI and family=Anzai, given=TOSHIHISA, given-i=TOSHIHISA and family=Inomata, given=TAKAYUKI, given-i=TAKAYUKI and family=Kinugawa, given=KOICHIRO, given-i=KOICHIRO and family=Sakata, given=YASUSHI, given-i=YASUSHI and family=Sato, given=NAOKI, given-i=NAOKI and family=Tsutsui, given=HIROYUKI, given-i=HIROYUKI and family=Yamamoto, given=KAZUHIRO, given-i=KAZUHIRO and family=Yoshimura, given=MICHIHIRO, given-i=MICHIHIRO and family=Saito, given=YOSHIHIKO, given-i=YOSHIHIKO and family=Kuwahara, given=KOICHIRO, given-i=KOICHIRO},
  date = {2025-09-01},
  journaltitle = {Journal of Cardiac Failure},
  shortjournal = {J. Card. Fail.},
  volume = {31},
  number = {9},
  pages = {1453--1459},
  issn = {1071-9164},
  doi = {10.1016/j.cardfail.2025.03.005},
  url = {https://www.sciencedirect.com/science/article/pii/S1071916425001393},
  urldate = {2026-07-30}
}

@article{rafieECGINTERP2021,
  title = {{{ECG Interpretation}}: {{Clinical Relevance}}, {{Challenges}}, and {{Advances}}},
  shorttitle = {{{ECG Interpretation}}},
  author = {Rafie, Nikita and Kashou, Anthony H. and Noseworthy, Peter A.},
  date = {2021-12},
  journaltitle = {Hearts},
  volume = {2},
  number = {4},
  pages = {505--513},
  publisher = {Multidisciplinary Digital Publishing Institute},
  issn = {2673-3846},
  doi = {10.3390/hearts2040039},
  url = {https://www.mdpi.com/2673-3846/2/4/39},
  urldate = {2026-07-30},
  langid = {english}
}

@article{sarkerDEEPLEARN2021,
  title = {Deep {{Learning}}: {{A Comprehensive Overview}} on {{Techniques}}, {{Taxonomy}}, {{Applications}} and {{Research Directions}}},
  shorttitle = {Deep {{Learning}}},
  author = {Sarker, Iqbal H.},
  date = {2021},
  journaltitle = {Sn Computer Science},
  shortjournal = {Sn Comput. Sci.},
  volume = {2},
  number = {6},
  eprint = {34426802},
  eprinttype = {pubmed},
  pages = {420},
  issn = {2662-995X},
  doi = {10.1007/s42979-021-00815-1},
  url = {https://pmc.ncbi.nlm.nih.gov/articles/PMC8372231/},
  urldate = {2026-07-30},
  pmcid = {PMC8372231}
}

@article{stahlschmidtMULTIMODAL2022,
  title = {Multimodal Deep Learning for Biomedical Data Fusion: A Review},
  shorttitle = {Multimodal Deep Learning for Biomedical Data Fusion},
  author = {Stahlschmidt, Sören Richard and Ulfenborg, Benjamin and Synnergren, Jane},
  date = {2022-03-01},
  journaltitle = {Briefings in Bioinformatics},
  shortjournal = {Brief. Bioinform.},
  volume = {23},
  number = {2},
  pages = {bbab569},
  issn = {1477-4054},
  doi = {10.1093/bib/bbab569},
  url = {https://doi.org/10.1093/bib/bbab569},
  urldate = {2026-07-31}
}

@article{huddarPREDICTING2016,
  title = {Predicting {{Complications}} in {{Critical Care Using Heterogeneous Clinical Data}}},
  author = {Huddar, Vijay and Desiraju, Bapu Koundinya and Rajan, Vaibhav and Bhattacharya, Sakyajit and Roy, Shourya and Reddy, Chandan K.},
  date = {2016},
  journaltitle = {IEEE Access},
  volume = {4},
  pages = {7988--8001},
  issn = {2169-3536},
  doi = {10.1109/ACCESS.2016.2618775},
  url = {https://ieeexplore.ieee.org/abstract/document/7593335},
  urldate = {2026-07-31}
}

@article{kronesREVIEWMUL2025,
  title = {Review of Multimodal Machine Learning Approaches in Healthcare},
  author = {Krones, Felix and Marikkar, Umar and Parsons, Guy and Szmul, Adam and Mahdi, Adam},
  date = {2025-02-01},
  journaltitle = {Information Fusion},
  shortjournal = {Inf. Fusion},
  volume = {114},
  pages = {102690},
  issn = {1566-2535},
  doi = {10.1016/j.inffus.2024.102690},
  url = {https://www.sciencedirect.com/science/article/pii/S1566253524004688},
  urldate = {2026-07-31}
}

@article{hannunCARDIOLOGI2019,
  title = {Cardiologist-Level Arrhythmia Detection and Classification in Ambulatory Electrocardiograms Using a Deep Neural Network},
  author = {Hannun, Awni Y. and Rajpurkar, Pranav and Haghpanahi, Masoumeh and Tison, Geoffrey H. and Bourn, Codie and Turakhia, Mintu P. and Ng, Andrew Y.},
  date = {2019-01},
  journaltitle = {Nature Medicine},
  shortjournal = {Nat. Med.},
  volume = {25},
  number = {1},
  pages = {65--69},
  publisher = {Nature Publishing Group},
  issn = {1546-170X},
  doi = {10.1038/s41591-018-0268-3},
  url = {https://www.nature.com/articles/s41591-018-0268-3},
  urldate = {2026-08-01},
  langid = {english}
}

@article{strodthoffDEEPLEARN2021,
  title = {Deep {{Learning}} for {{ECG Analysis}}: {{Benchmarks}} and {{Insights}} from {{PTB-XL}}},
  shorttitle = {Deep {{Learning}} for {{ECG Analysis}}},
  author = {Strodthoff, Nils and Wagner, Patrick and Schaeffter, Tobias and Samek, Wojciech},
  date = {2021-05},
  journaltitle = {IEEE Journal of Biomedical and Health Informatics},
  shortjournal = {IEEE J. Biomed. Health Inform.},
  volume = {25},
  number = {5},
  pages = {1519--1528},
  issn = {2168-2208},
  doi = {10.1109/JBHI.2020.3022989},
  url = {https://ieeexplore.ieee.org/abstract/document/9190034},
  urldate = {2026-08-01}
}

@article{yuAIBASEDST2022,
  title = {{{AI-Based Stroke Disease Prediction System Using ECG}} and {{PPG Bio-Signals}}},
  author = {Yu, Jaehak and Park, Sejin and Kwon, Soon-Hyun and Cho, Kang-Hee and Lee, Hansung},
  date = {2022},
  journaltitle = {IEEE Access},
  volume = {10},
  pages = {43623--43638},
  issn = {2169-3536},
  doi = {10.1109/ACCESS.2022.3169284},
  url = {https://ieeexplore.ieee.org/abstract/document/9761215/keywords},
  urldate = {2026-08-01}
}

@article{rathnayakePHYSIOLOGI2026,
  title = {Physiological {{Monitoring Applications}} of {{Wearable Multimodal Fusion Systems Based}} on {{ECG}} and {{PPG}}: {{A Comprehensive Review}}},
  shorttitle = {Physiological {{Monitoring Applications}} of {{Wearable Multimodal Fusion Systems Based}} on {{ECG}} and {{PPG}}},
  author = {Rathnayake, Chamod and Chen, Wenjing and Jayawickrama, Sahan and Lai, Dakun},
  date = {2026-01},
  journaltitle = {Sensors},
  volume = {26},
  number = {11},
  pages = {3477},
  publisher = {Multidisciplinary Digital Publishing Institute},
  issn = {1424-8220},
  doi = {10.3390/s26113477},
  url = {https://www.mdpi.com/1424-8220/26/11/3477},
  urldate = {2026-08-01},
  langid = {english}
}

@article{xiaoDEEPLEARN2023,
  title = {Deep {{Learning-Based ECG Arrhythmia Classification}}: {{A Systematic Review}}},
  shorttitle = {Deep {{Learning-Based ECG Arrhythmia Classification}}},
  author = {Xiao, Qiao and Lee, Khuan and Mokhtar, Siti Aisah and Ismail, Iskasymar and Pauzi, Ahmad Luqman bin Md and Zhang, Qiuxia and Lim, Poh Ying},
  date = {2023-01},
  journaltitle = {Applied Sciences},
  shortjournal = {Appl. Sci.},
  volume = {13},
  number = {8},
  pages = {4964},
  publisher = {Multidisciplinary Digital Publishing Institute},
  issn = {2076-3417},
  doi = {10.3390/app13084964},
  url = {https://www.mdpi.com/2076-3417/13/8/4964},
  urldate = {2026-08-01},
  langid = {english}
}

@article{teymouriECGFREQUE2022,
  title = {{{ECG}} Frequency Changes in Potassium Disorders: A Narrative Review},
  shorttitle = {{{ECG}} Frequency Changes in Potassium Disorders},
  author = {Teymouri, Navid and Mesbah, Sahar and Navabian, Seyed Mohammad Hossein and Shekouh, Dorsa and Najafabadi, Mahsa Mohammadi and Norouzkhani, Narges and Poudineh, Mohadeseh and Qadirifard, Mohammad Sadegh and Mehrtabar, Saba and Deravi, Niloofar},
  date = {2022-06-15},
  journaltitle = {American Journal of Cardiovascular Disease},
  shortjournal = {Am. J. Cardiovasc. Dis.},
  volume = {12},
  number = {3},
  eprint = {35873184},
  eprinttype = {pubmed},
  pages = {112--124},
  issn = {2160-200X},
  url = {https://pmc.ncbi.nlm.nih.gov/articles/PMC9301030/},
  urldate = {2026-08-02},
  pmcid = {PMC9301030}
}

@article{arnsdorfELECTROCAR1976,
  title = {Electrocardiogram in {{Hyperkalemia}}: {{Electrocardiographic Pattern}} of {{Anteroseptal Myocardial Infarction Mimicked}} by {{Hyperkalemia-Induced Disturbance}} of {{Impulse Conduction}}},
  shorttitle = {Electrocardiogram in {{Hyperkalemia}}},
  author = {Arnsdorf, Morton F.},
  date = {1976-10-01},
  journaltitle = {Archives of Internal Medicine},
  shortjournal = {Arch. Intern. Med.},
  volume = {136},
  number = {10},
  pages = {1161--1163},
  issn = {0003-9926},
  doi = {10.1001/archinte.1976.03630100073019},
  url = {https://doi.org/10.1001/archinte.1976.03630100073019},
  urldate = {2026-08-02}
}

@article{ettingerHYPERKALEM1974,
  title = {Hyperkalemia, Cardiac Conduction, and the Electrocardiogram: {{A}} Review},
  shorttitle = {Hyperkalemia, Cardiac Conduction, and the Electrocardiogram},
  author = {Ettinger, Philip O. and Regan, Timothy J. and Oldewurtel, Henry A.},
  date = {1974-09-01},
  journaltitle = {American Heart Journal},
  shortjournal = {Am. Heart J.},
  volume = {88},
  number = {3},
  pages = {360--371},
  issn = {0002-8703},
  doi = {10.1016/0002-8703(74)90473-6},
  url = {https://www.sciencedirect.com/science/article/pii/0002870374904736},
  urldate = {2026-08-02}
}

@online{OVERVIEWC2018,
  title = {Overview | {{Chronic}} Heart Failure in Adults: Diagnosis and Management | {{Guidance}} | {{NICE}}},
  shorttitle = {Overview | {{Chronic}} Heart Failure in Adults},
  date = {2018-09-12},
  publisher = {NICE},
  url = {https://www.nice.org.uk/guidance/ng106},
  urldate = {2026-08-03},
  langid = {english}
}

@article{PhysioNet-mimiciv-3.1,
  author = {Johnson, Alistair and Bulgarelli, Lucas and Pollard, Tom and Gow, Brian and Moody, Benjamin and Horng, Steven and Celi, Leo Anthony and Mark, Roger},
  title = {{MIMIC-IV}},
  journal = {{PhysioNet}},
  year = {2024},
  month = oct,
  note = {Version 3.1},
  doi = {10.13026/kpb9-mt58},
  url = {https://doi.org/10.13026/kpb9-mt58}
}

@article{johnsonMIMICIVFR2023,
  title = {{{MIMIC-IV}}, a Freely Accessible Electronic Health Record Dataset},
  author = {Johnson, Alistair E. W. and Bulgarelli, Lucas and Shen, Lu and Gayles, Alvin and Shammout, Ayad and Horng, Steven and Pollard, Tom J. and Hao, Sicheng and Moody, Benjamin and Gow, Brian and Lehman, Li-wei H. and Celi, Leo A. and Mark, Roger G.},
  date = {2023-01-03},
  journaltitle = {Scientific Data},
  shortjournal = {Sci. Data},
  volume = {10},
  number = {1},
  pages = {1},
  publisher = {Nature Publishing Group},
  issn = {2052-4463},
  doi = {10.1038/s41597-022-01899-x},
  url = {https://www.nature.com/articles/s41597-022-01899-x},
  urldate = {2026-08-03},
  langid = {english}
}

@article{pollardPHYSIONET2026,
  title = {{{PhysioNet}} as a Global Platform for Biomedical Research},
  author = {Pollard, Tom and Moody, Benjamin E. and Lehman, Li-wei H. and Gow, Brian J. and Fernandes, Chrystinne and Xie, Chen and Johnson, Alistair and Mark, Roger G. and Heldt, Thomas},
  date = {2026-03-26},
  journaltitle = {Nature Health},
  shortjournal = {Nat. Health},
  volume = {1},
  number = {8},
  pages = {792--795},
  issn = {3005-0693},
  doi = {10.1038/s44360-026-00096-z},
  url = {https://www.nature.com/articles/s44360-026-00096-z},
  urldate = {2026-08-03},
  langid = {english}
}

@dataset{gowMIMICIVECG,
  title = {{{MIMIC-IV-ECG}}: {{Diagnostic Electrocardiogram Matched Subset}}},
  shorttitle = {{{MIMIC-IV-ECG}}},
  author = {Gow, Brian and Pollard, Tom and Nathanson, Larry A and Johnson, Alistair and Moody, Benjamin and Fernandes, Chrystinne and Greenbaum, Nathaniel and Waks, Jonathan W and Eslami, Parastou and Carbonati, Tanner and Chaudhari, Ashish and Herbst, Elizabeth and Moukheiber, Dana and Berkowitz, Seth and Mark, Roger and Horng, Steven},
  publisher = {PhysioNet},
  doi = {10.13026/4NQG-SB35},
  url = {https://physionet.org/content/mimic-iv-ecg/1.0/},
  urldate = {2026-08-03},
  version = {1.0}
}

@dataset{johnsonMIMICIVED,
  title = {{{MIMIC-IV-ED}}},
  author = {Johnson, Alistair and Bulgarelli, Lucas and Pollard, Tom and Celi, Leo Anthony and Mark, Roger and Horng, Steven},
  publisher = {PhysioNet},
  doi = {10.13026/5NTK-KM72},
  url = {https://physionet.org/content/mimic-iv-ed/2.2/},
  urldate = {2026-08-03},
  version = {2.2}
}

@online{MIMICIV,
  title = {{{MIMIC-IV}}},
  url = {https://mimic.mit.edu/docs/iv/},
  urldate = {2026-08-04},
  langid = {american},
  organization = {MIMIC}
}

@online{alcarazMULTIMODAL2026,
  title = {A {{Multimodal Deep Learning Framework}} for {{Predicting ICU Deterioration}}: {{Integrating ECG Waveforms}} with {{Clinical Data}} and {{Clinician Benchmarking}}},
  shorttitle = {A {{Multimodal Deep Learning Framework}} for {{Predicting ICU Deterioration}}},
  author = {Alcaraz, Juan Miguel López and Moran, Xicoténcatl López and Zaragoza, Erick Dávila and Händel, Claas and Koebe, Richard and Haverkamp, Wilhelm and Strodthoff, Nils},
  date = {2026-01-10},
  eprint = {2601.06645},
  eprinttype = {arXiv},
  eprintclass = {eess.SP},
  doi = {10.48550/arXiv.2601.06645},
  url = {http://arxiv.org/abs/2601.06645},
  urldate = {2026-08-04},
  pubstate = {prepublished}
}

@inproceedings{seabold2010statsmodels,
  title={statsmodels: Econometric and statistical modeling with python},
  author={Seabold, Skipper and Perktold, Josef},
  booktitle={9th Python in Science Conference},
  year={2010},
}

@article{scikit-learn,
  title={Scikit-learn: Machine Learning in {P}ython},
  author={Pedregosa, F. and Varoquaux, G. and Gramfort, A. and Michel, V.
          and Thirion, B. and Grisel, O. and Blondel, M. and Prettenhofer, P.
          and Weiss, R. and Dubourg, V. and Vanderplas, J. and Passos, A. and
          Cournapeau, D. and Brucher, M. and Perrot, M. and Duchesnay, E.},
  journal={Journal of Machine Learning Research},
  volume={12},
  pages={2825--2830},
  year={2011}
}

@article{teramotoLONGITUDIN2024,
    title = {Longitudinal {NT}‐{proBNP}: {Associations} {With} {Echocardiographic} {Changes} and {Outcomes} in {Heart} {Failure}},
    volume = {13},
    shorttitle = {Longitudinal {NT}‐{proBNP}},
    url = {https://www.ahajournals.org/doi/full/10.1161/JAHA.123.032254},
    doi = {10.1161/JAHA.123.032254},
    number = {9},
    urldate = {2026-08-11},
    journal = {Journal of the American Heart Association},
    publisher = {Wiley},
    author = {Teramoto, Kanako and Tay, Wan Ting and Tromp, Jasper and Ouwerkerk, Wouter and Teng, Tiew‐Hwa Katherine and Chandramouli, Chanchal and Liew, Oi Wah and Chong, Jenny and Poppe, Katrina K. and Lund, Mayanna and Devlin, Gerry and Troughton, Richard W. and Doughty, Robert N. and Richards, Arthur Mark and Lam, Carolyn S. P.},
    month = may,
    year = {2024},
    pages = {e032254},
}

@article{bayes-genisPRACTICAL2023,
    title = {Practical {Algorithms} for {Early} {Diagnosis} of {Heart} {Failure} and {Heart} {Stress} {Using} {NT}-{proBNP}: {A} {Clinical} {Consensus} {Statement} from the {Heart} {Failure} {Association} of the {ESC}},
    volume = {25},
    issn = {1388-9842},
    shorttitle = {Practical {Algorithms} for {Early} {Diagnosis} of {Heart} {Failure} and {Heart} {Stress} {Using} {NT}-{proBNP}},
    url = {https://doi.org/10.1002/ejhf.3036},
    doi = {10.1002/ejhf.3036},
    number = {11},
    urldate = {2026-08-11},
    journal = {European Journal of Heart Failure},
    author = {Bayes-Genis, Antoni and Docherty, Kieran F. and Petrie, Mark C. and Januzzi, James L. and Mueller, Christian and Anderson, Lisa and Bozkurt, Biykem and Butler, Javed and Chioncel, Ovidiu and Cleland, John G.F. and Christodorescu, Ruxandra and Del Prato, Stefano and Gustafsson, Finn and Lam, Carolyn S.P. and Moura, Brenda and Pop-Busui, Rodica and Seferovic, Petar and Volterrani, Maurizio and Vaduganathan, Muthiah and Metra, Marco and Rosano, Giuseppe},
    month = nov,
    year = {2023},
    pages = {1891--1898},
}

@article{magesELECTRICAL2021,
    title = {Electrical {Ventricular} {Remodeling} in {Dilated} {Cardiomyopathy}},
    volume = {10},
    copyright = {http://creativecommons.org/licenses/by/3.0/},
    issn = {2073-4409},
    url = {https://www.mdpi.com/2073-4409/10/10/2767},
    doi = {10.3390/cells10102767},
    language = {en},
    number = {10},
    urldate = {2026-08-11},
    journal = {Cells},
    publisher = {Multidisciplinary Digital Publishing Institute},
    author = {Mages, Christine and Gampp, Heike and Syren, Pascal and Rahm, Ann-Kathrin and André, Florian and Frey, Norbert and Lugenbiel, Patrick and Thomas, Dierk},
    month = oct,
    year = {2021},
    pages = {2767},
}

@article{hombachELECTROCAR2006,
    title = {Electrocardiography of the {Failing} {Heart}},
    volume = {24},
    issn = {0733-8651, 1558-2264},
    url = {https://www.cardiology.theclinics.com/article/S0733-8651(06)00011-7/abstract},
    doi = {10.1016/j.ccl.2006.03.002},
    language = {English},
    number = {3},
    urldate = {2026-08-11},
    journal = {Cardiology Clinics},
    publisher = {Elsevier},
    author = {Hombach, Vinzenz},
    month = aug,
    year = {2006},
    pages = {413--426},
}

@article{buxtonACCAHAHR2006,
    title = {{ACC}/{AHA}/{HRS} 2006 {Key} {Data} {Elements} and {Definitions} for {Electrophysiological} {Studies} and {Procedures}},
    volume = {48},
    url = {https://www.jacc.org/doi/abs/10.1016/j.jacc.2006.09.020},
    doi = {10.1016/j.jacc.2006.09.020},
    number = {11},
    urldate = {2026-08-16},
    journal = {JACC},
    publisher = {American College of Cardiology Foundation},
    author = {Buxton, Alfred E. and Calkins, Hugh and Callans, David J. and DiMarco, John P. and Fisher, John D. and Leon Greene, H. and Haines, David E. and Hayes, David L. and Heidenreich, Paul A. and Miller, John M. and Poppas, Athena and Prystowsky, Eric N. and Schoenfeld, Mark H. and Zimetbaum, Peter J. and Heidenreich, Paul A. and Goff, David C. and Grover, Frederick L. and Malenka, David J. and Peterson, Eric D. and Radford, Martha J. and Redberg, Rita F.},
    month = dec,
    year = {2006},
    pages = {2360--2396},
}

@article{surawiczAHAACCFH2009,
    title = {{AHA}/{ACCF}/{HRS} {Recommendations} for the {Standardization} and {Interpretation} of the {Electrocardiogram}: {Part} {III}: {Intraventricular} {Conduction} {Disturbances}: {A} {Scientific} {Statement} {From} the {American} {Heart} {Association} {Electrocardiography} and {Arrhythmias} {Committee}, {Council} on {Clinical} {Cardiology}; the {American} {College} of {Cardiology} {Foundation}; and the {Heart} {Rhythm} {Society}: \textit{{Endorsed} by the {International} {Society} for {Computerized} {Electrocardiology}}},
    volume = {119},
    issn = {0009-7322, 1524-4539},
    shorttitle = {{AHA}/{ACCF}/{HRS} {Recommendations} for the {Standardization} and {Interpretation} of the {Electrocardiogram}},
    url = {https://www.ahajournals.org/doi/10.1161/CIRCULATIONAHA.108.191095},
    doi = {10.1161/CIRCULATIONAHA.108.191095},
    language = {en},
    number = {10},
    urldate = {2026-08-16},
    journal = {Circulation},
    author = {Surawicz, Borys and Childers, Rory and Deal, Barbara J. and Gettes, Leonard S.},
    month = mar,
    year = {2009},
}

@article{kligfieldRECOMMENDA2007,
    title = {Recommendations for the {Standardization} and {Interpretation} of the {Electrocardiogram}: {Part} {I}: {The} {Electrocardiogram} and {Its} {Technology}: {A} {Scientific} {Statement} {From} the {American} {Heart} {Association} {Electrocardiography} and {Arrhythmias} {Committee}, {Council} on {Clinical} {Cardiology}; the {American} {College} of {Cardiology} {Foundation}; and the {Heart} {Rhythm} {Society} \textit{{Endorsed} by the {International} {Society} for {Computerized} {Electrocardiology}}},
    volume = {115},
    issn = {0009-7322, 1524-4539},
    shorttitle = {Recommendations for the {Standardization} and {Interpretation} of the {Electrocardiogram}},
    url = {https://www.ahajournals.org/doi/10.1161/CIRCULATIONAHA.106.180200},
    doi = {10.1161/CIRCULATIONAHA.106.180200},
    language = {en},
    number = {10},
    urldate = {2026-08-16},
    journal = {Circulation},
    author = {Kligfield, Paul and Gettes, Leonard S. and Bailey, James J. and Childers, Rory and Deal, Barbara J. and Hancock, E. William and Van Herpen, Gerard and Kors, Jan A. and Macfarlane, Peter and Mirvis, David M. and Pahlm, Olle and Rautaharju, Pentti and Wagner, Galen S.},
    month = mar,
    year = {2007},
    pages = {1306--1324},
}

@article{issaHEARTBEAT2023,
    title = {Heartbeat classification based on single lead-{II} {ECG} using deep learning},
    volume = {9},
    issn = {2405-8440},
    url = {https://www.cell.com/heliyon/abstract/S2405-8440(23)05182-4},
    doi = {10.1016/j.heliyon.2023.e17974},
    language = {English},
    number = {7},
    urldate = {2026-08-16},
    journal = {Heliyon},
    publisher = {Elsevier},
    author = {Issa, Mohamed F. and Yousry, Ahmed and Tuboly, Gergely and Juhasz, Zoltan and AbuEl-Atta, Ahmed H. and Selim, Mazen M.},
    month = jul,
    year = {2023},
}

@article{baggishDIFFERENTI2008,
    series = {An {International} {Consensus} {Statement} {Regarding} {Amino}-{Terminal} {Pro}-{B}-{Type} {Natriuretic} {Peptide} {Testing}: {The} {International} {NT}-{proBNP} {Consensus} {Panel}},
    title = {The {Differential} {Diagnosis} of an {Elevated} {Amino}-{Terminal} {Pro}–{B}-{Type} {Natriuretic} {Peptide} {Level}},
    volume = {101},
    issn = {0002-9149},
    url = {https://www.sciencedirect.com/science/article/pii/S0002914907022448},
    doi = {10.1016/j.amjcard.2007.11.019},
    number = {3, Supplement},
    urldate = {2026-08-16},
    journal = {The American Journal of Cardiology},
    author = {Baggish, Aaron L. and van Kimmenade, Roland R. J. and Januzzi, James L.},
    month = feb,
    year = {2008},
    pages = {S43--S48},
}

@article{COMPARISON,
    title = {Comparison of {BNP} and {NT}-{proBNP} in {Patients} {With} {Heart} {Failure} and {Reduced} {Ejection} {Fraction}},
    url = {https://www.ahajournals.org/doi/10.1161/CIRCHEARTFAILURE.119.006541},
    language = {en},
    urldate = {2026-08-16},
    journal = {Circulation: Heart Failure},
}

@misc{guEFFICIENTL2022,
    title = {Efficiently {Modeling} {Long} {Sequences} with {Structured} {State} {Spaces}},
    url = {http://arxiv.org/abs/2111.00396},
    doi = {10.48550/arXiv.2111.00396},
    urldate = {2026-08-17},
    publisher = {arXiv},
    author = {Gu, Albert and Goel, Karan and Ré, Christopher},
    month = aug,
    year = {2022},
    note = {arXiv:2111.00396 [cs.LG]},
}

@misc{mehariADVANCING2022,
    title = {Advancing the {State}-of-the-{Art} for {ECG} {Analysis} through {Structured} {State} {Space} {Models}},
    url = {http://arxiv.org/abs/2211.07579},
    doi = {10.48550/arXiv.2211.07579},
    urldate = {2026-08-17},
    publisher = {arXiv},
    author = {Mehari, Temesgen and Strodthoff, Nils},
    month = nov,
    year = {2022},
    note = {arXiv:2211.07579 [cs.LG]},
}

@inproceedings{holzmullerBETTERDEF2024,
    address = {Vancouver (BC), Canada},
    title = {Better by {Default}: {Strong} {Pre}-{Tuned} {MLPs} and {Boosted} {Trees} on {Tabular} {Data}},
    shorttitle = {Better by {Default}},
    url = {https://hal.science/hal-04641923},
    doi = {10.48550/arXiv.2407.04491},
    urldate = {2026-08-17},
    booktitle = {Neural {Information} {Processing} {Systems}},
    author = {Holzmüller, David and Grinsztajn, Léo and Steinwart, Ingo},
    month = dec,
    year = {2024},
}

@article{januzziNTPROBNPT2006,
    title = {{NT}-{proBNP} testing for diagnosis and short-term prognosis in acute destabilized heart failure: an international pooled analysis of 1256 patients: {The} {International} {Collaborative} of {NT}-{proBNP} {Study}},
    volume = {27},
    issn = {0195-668X},
    shorttitle = {{NT}-{proBNP} testing for diagnosis and short-term prognosis in acute destabilized heart failure},
    url = {https://doi.org/10.1093/eurheartj/ehi631},
    doi = {10.1093/eurheartj/ehi631},
    number = {3},
    urldate = {2026-08-20},
    journal = {European Heart Journal},
    author = {Januzzi, James L. and van Kimmenade, Roland and Lainchbury, John and Bayes-Genis, Antoni and Ordonez-Llanos, Jordi and Santalo-Bel, Miguel and Pinto, Yigal M. and Richards, Mark},
    month = feb,
    year = {2006},
    pages = {330--337},
}

@article{heidenreich2022AHAA2022,
    title = {2022 {AHA}/{ACC}/{HFSA} {Guideline} for the {Management} of {Heart} {Failure}: {A} {Report} of the {American} {College} of {Cardiology}/{American} {Heart} {Association} {Joint} {Committee} on {Clinical} {Practice} {Guidelines}},
    volume = {145},
    issn = {0009-7322, 1524-4539},
    shorttitle = {2022 {AHA}/{ACC}/{HFSA} {Guideline} for the {Management} of {Heart} {Failure}},
    url = {https://www.ahajournals.org/doi/10.1161/CIR.0000000000001063},
    doi = {10.1161/CIR.0000000000001063},
    language = {en},
    number = {18},
    urldate = {2026-08-20},
    journal = {Circulation},
    author = {Heidenreich, Paul A. and Bozkurt, Biykem and Aguilar, David and Allen, Larry A. and Byun, Joni J. and Colvin, Monica M. and Deswal, Anita and Drazner, Mark H. and Dunlay, Shannon M. and Evers, Linda R. and Fang, James C. and Fedson, Savitri E. and Fonarow, Gregg C. and Hayek, Salim S. and Hernandez, Adrian F. and Khazanie, Prateeti and Kittleson, Michelle M. and Lee, Christopher S. and Link, Mark S. and Milano, Carmelo A. and Nnacheta, Lorraine C. and Sandhu, Alexander T. and Stevenson, Lynne Warner and Vardeny, Orly and Vest, Amanda R. and Yancy, Clyde W.},
    month = may,
    year = {2022},
}

@article{kimPOTASSIUM,
    title = {Potassium {Disorders}: {Hypokalemia} and {Hyperkalemia}},
    language = {en},
    author = {Kim, Michael J and Valerio, Christina and Knobloch, Glynnis K},
}

@article{battelinoCLINICALT2019,
    title = {Clinical {Targets} for {Continuous} {Glucose} {Monitoring} {Data} {Interpretation}: {Recommendations} {From} the {International} {Consensus} on {Time} in {Range}},
    volume = {42},
    issn = {0149-5992},
    shorttitle = {Clinical {Targets} for {Continuous} {Glucose} {Monitoring} {Data} {Interpretation}},
    url = {https://doi.org/10.2337/dci19-0028},
    doi = {10.2337/dci19-0028},
    number = {8},
    urldate = {2026-08-22},
    journal = {Diabetes Care},
    author = {Battelino, Tadej and Danne, Thomas and Bergenstal, Richard M. and Amiel, Stephanie A. and Beck, Roy and Biester, Torben and Bosi, Emanuele and Buckingham, Bruce A. and Cefalu, William T. and Close, Kelly L. and Cobelli, Claudio and Dassau, Eyal and DeVries, J. Hans and Donaghue, Kim C. and Dovc, Klemen and Doyle, III, Francis J. and Garg, Satish and Grunberger, George and Heller, Simon and Heinemann, Lutz and Hirsch, Irl B. and Hovorka, Roman and Jia, Weiping and Kordonouri, Olga and Kovatchev, Boris and Kowalski, Aaron and Laffel, Lori and Levine, Brian and Mayorov, Alexander and Mathieu, Chantal and Murphy, Helen R. and Nimri, Revital and Nørgaard, Kirsten and Parkin, Christopher G. and Renard, Eric and Rodbard, David and Saboo, Banshi and Schatz, Desmond and Stoner, Keaton and Urakami, Tatsuiko and Weinzimer, Stuart A. and Phillip, Moshe},
    month = jun,
    year = {2019},
    pages = {1593--1603},
}

@article{ferraroLABORATORY2016,
    title = {Laboratory medicine in the new healthcare environment},
    volume = {54},
    issn = {1437-4331, 1434-6621},
    url = {https://www.degruyter.com/document/doi/10.1515/cclm-2015-0803/html},
    doi = {10.1515/cclm-2015-0803},
    number = {4},
    urldate = {2026-08-25},
    journal = {Clinical Chemistry and Laboratory Medicine (CCLM)},
    author = {Ferraro, Simona and Braga, Federica and Panteghini, Mauro},
    month = jan,
    year = {2016},
}

@article{mckieNTPROBNP2016,
    title = {{NT}-{proBNP}},
    volume = {68},
    issn = {07351097},
    url = {https://linkinghub.elsevier.com/retrieve/pii/S0735109716364622},
    doi = {10.1016/j.jacc.2016.10.001},
    language = {en},
    number = {22},
    urldate = {2026-08-25},
    journal = {Journal of the American College of Cardiology},
    author = {McKie, Paul M. and Burnett, John C.},
    month = dec,
    year = {2016},
    pages = {2437--2439},
}

@article{yeshouaINTERVENTI2023,
    title = {Interventions to reduce repetitive ordering of low-value inpatient laboratory tests: a systematic review},
    volume = {12},
    issn = {2399-6641},
    shorttitle = {Interventions to reduce repetitive ordering of low-value inpatient laboratory tests},
    url = {https://qir.bmj.com/lookup/doi/10.1136/bmjoq-2022-002128},
    doi = {10.1136/bmjoq-2022-002128},
    language = {en},
    number = {1},
    urldate = {2026-08-25},
    journal = {BMJ Open Quality},
    author = {Yeshoua, Brandon and Bowman, Chip and Dullea, Jonathan and Ditkowsky, Jared and Shyu, Margaret and Lam, Hansen and Zhao, William and Shin, Joo Yeon and Dunn, Andrew and Tsega, Surafel and S. Linker, Anne and Shah, Manan},
    month = mar,
    year = {2023},
    pages = {e002128},
}

@article{lima-oliveiraLABORATORY2015,
    title = {Laboratory {Diagnostics} and {Quality} of {Blood} {Collection}},
    volume = {34},
    issn = {1452-8258},
    url = {https://pmc.ncbi.nlm.nih.gov/articles/PMC4922344/},
    doi = {10.2478/jomb-2014-0043},
    number = {3},
    urldate = {2026-08-25},
    journal = {Journal of Medical Biochemistry},
    author = {Lima-Oliveira, Gabriel and Lippi, Giuseppe and Salvagno, Gian Luca and Picheth, Geraldo and Guidi, Gian Cesare},
    month = jul,
    year = {2015},
    pages = {288--294},
}

@article{petersenREGULATION2017,
    title = {Regulation of hepatic glucose metabolism in health and disease},
    volume = {13},
    copyright = {2017 Springer Nature Limited},
    issn = {1759-5037},
    url = {https://www.nature.com/articles/nrendo.2017.80},
    doi = {10.1038/nrendo.2017.80},
    language = {en},
    number = {10},
    urldate = {2026-08-25},
    journal = {Nature Reviews Endocrinology},
    publisher = {Nature Publishing Group},
    author = {Petersen, Max C. and Vatner, Daniel F. and Shulman, Gerald I.},
    month = oct,
    year = {2017},
    pages = {572--587},
}

@article{indolfiDEVELOPMEN2026,
    title = {Development and {External} {Validation} of an {AI}-{ECG} {Algorithm} for {Estimating} {Elevated} {Serum} {NT}-{proBNP} {Levels}},
    issn = {2058-1742},
    url = {https://doi.org/10.1093/ehjqcco/qcag128},
    doi = {10.1093/ehjqcco/qcag128},
    urldate = {2026-08-25},
    journal = {European Heart Journal - Quality of Care and Clinical Outcomes},
    author = {Indolfi, Ciro and Spaccarotella, Carmen and Polimeni, Alberto and Castiello, Domenico Simone and Esposito, Giovanni and Curcio, Antonio and Gravina, Adriana and Leone, Nicola and Kim, Joonghee and Garoia, Federico and Sinagra, Gianfranco and Lüscher, Thomas F and Cho, Youngjin},
    month = aug,
    year = {2026},
    pages = {qcag128},
}

@article{abdinEFFECTSDA2024,
    title = {Effects of {Dapagliflozin} {According} to {QRS} {Duration} {Across} the {Spectrum} of {Left} {Ventricular} {Ejection} {Fraction}: {An} {Analysis} of {DAPA}-{HF} and {DELIVER}},
    volume = {26},
    issn = {1388-9842},
    shorttitle = {Effects of {Dapagliflozin} {According} to {QRS} {Duration} {Across} the {Spectrum} of {Left} {Ventricular} {Ejection} {Fraction}},
    url = {https://doi.org/10.1002/ejhf.3350},
    doi = {10.1002/ejhf.3350},
    number = {9},
    urldate = {2026-08-25},
    journal = {European Journal of Heart Failure},
    author = {Abdin, Amr and Kondo, Toru and Böhm, Michael and Jhund, Pardeep S. and Claggett, Brian L. and Vaduganathan, Muthiah and Hernandez, Adrian F. and Lam, Carolyn S.P. and Inzucchi, Silvio E. and Martinez, Felipe A. and de Boer, Rudolf A. and Desai, Akshay S. and Køber, Lars and Sabatine, Marc S. and Petersson, Magnus and Bachus, Erasmus and Solomon, Scott D. and McMurray, John J.V. and {the DAPA-HF and DELIVER Committees and Investigators}},
    month = sep,
    year = {2024},
    pages = {1952--1963},
}

@article{neyaziDEEPLEARN2024,
    title = {Deep learning-based {NT}-{proBNP} prediction from the {ECG} for risk assessment in the community},
    volume = {62},
    copyright = {De Gruyter expressly reserves the right to use all content for commercial text and data mining within the meaning of Section 44b of the German Copyright Act.},
    issn = {1437-4331},
    url = {https://www.degruyterbrill.com/document/doi/10.1515/cclm-2023-0743/html?lang=en},
    doi = {10.1515/cclm-2023-0743},
    language = {en},
    number = {4},
    urldate = {2026-08-25},
    journal = {Clinical Chemistry and Laboratory Medicine (CCLM)},
    publisher = {De Gruyter},
    author = {Neyazi, Meraj and Bremer, Jan P. and Knorr, Marius S. and Gross, Stefan and Brederecke, Jan and Schweingruber, Nils and Csengeri, Dora and Schrage, Benedikt and Bahls, Martin and Friedrich, Nele and Zeller, Tanja and Felix, Stephan and Blankenberg, Stefan and Dörr, Marcus and Vollmer, Marcus and Schnabel, Renate B.},
    month = mar,
    year = {2024},
    pages = {740--752},
}

@article{al-masudBENCHMARKI2026,
    title = {Benchmarking {ECG} {FMs}: {A} {Reality} {Check} {Across} {Clinical} {Tasks}},
    volume = {2026},
    shorttitle = {Benchmarking {ECG} {FMs}},
    url = {https://proceedings.iclr.cc/paper_files/paper/2026/hash/ed49dc4569d0c93dab63c6e0a045f6c5-Abstract-Conference.html},
    language = {en},
    urldate = {2026-08-25},
    journal = {International Conference on Learning Representations},
    author = {Al-Masud, M. A. and Lopez Alcaraz, Juan Miguel and Strodthoff, Nils},
    month = apr,
    year = {2026},
    pages = {146689--146715},
}

@misc{wijkACUTELINES2026,
    title = {Acutelines: a {Dutch} emergency department-based data-biobank for multimodal, longitudinal acute care research},
    copyright = {© 2026, Posted by openRxiv. This pre-print is available under a Creative Commons License (Attribution 4.0 International), CC BY 4.0, as described at http://creativecommons.org/licenses/by/4.0/},
    shorttitle = {Acutelines},
    url = {https://www.medrxiv.org/content/10.64898/2026.07.30.26359325v1},
    doi = {10.64898/2026.07.30.26359325},
    language = {en},
    urldate = {2026-08-25},
    publisher = {medRxiv},
    author = {Wijk, Raymond J. van and Alcaraz, Juan Miguel Lopez and Schoonhoven, Anna D. and Li, Jie and Horst, Sanne Ter and Voert, Michèle A. ter and Eerens, Annet M. and Bastide, Anna J. la and Postema, Jochem and Maaten, Jan C. ter and Avest, Ewoud ter and Strodthoff, Nils and Bouma, Hjalmar R.},
    month = aug,
    year = {2026},
    note = {ISSN: 3067-2007
Pages: 2026.07.30.26359325},
}

@online{vanwijkECGSPECTR2026,
  title = {{{ECG}} Spectrogram-Based Deep Learning Model to Predict Deterioration of Patients with Early Sepsis at the Emergency Department: A Study from the {{Acutelines}} Data- and Biobank},
  shorttitle = {{{ECG}} Spectrogram-Based Deep Learning Model to Predict Deterioration of Patients with Early Sepsis at the Emergency Department},
  author = {Van Wijk, Raymond J. and Schoonhoven, Anna D. and De Vree, Leanne and Ter Horst, Sanne and Gaidhane, Chetan and Alcaraz, Juan Miguel Lopez and Strodthoff, Nils and Ter Maaten, Jan C. and Bouma, Hjalmar R. and Li, Jie},
  date = {2026-03-27},
  eprinttype = {Emergency Medicine},
  doi = {10.64898/2026.03.26.26349371},
  url = {http://medrxiv.org/lookup/doi/10.64898/2026.03.26.26349371},
  urldate = {2026-08-25},
  langid = {english},
  pubstate = {prepublished}
}

@inproceedings{lopezalcarazEXPLAINING2025,
    title = {Explaining {Time} {Series} {Classification} {Predictions} via {Causal} {Attributions}},
    issn = {2375-0197},
    url = {https://ieeexplore.ieee.org/abstract/document/11272614},
    doi = {10.1109/ICTAI66417.2025.00014},
    urldate = {2026-08-25},
    booktitle = {2025 {IEEE} 37th {International} {Conference} on {Tools} with {Artificial} {Intelligence} ({ICTAI})},
    author = {Lopez Alcaraz, Juan Miguel and Strodthoff, Nils},
    month = nov,
    year = {2025},
    note = {ISSN: 2375-0197},
    pages = {42--51},
}

\end{document}